\documentclass[pdflatex,sn-vancouver-num,iicol]{sn-jnl}

\newcommand{\ttitle}{PEBRE: An Open-Hardware Compute and Perception Add-On for the Pepper Robot}
\newcommand\addonname{PEBRE}

\usepackage{graphicx}%
\usepackage{multirow}
\usepackage{amsmath,amssymb,amsfonts}%
\usepackage{amsthm}%
\usepackage{mathrsfs}%
\usepackage[title]{appendix}%
\usepackage{xcolor}%
\usepackage{textcomp}%
\usepackage{manyfoot}%
\usepackage{algorithm}%
\usepackage{algorithmicx}%
\usepackage{algpseudocode}%
\usepackage{listings}%

\usepackage[T1]{fontenc}
\usepackage{lmodern}
\usepackage[colorinlistoftodos,textsize=footnotesize,textwidth=\marginparwidth,%
backgroundcolor=orange!30]{todonotes}%
\definecolor{nng}{RGB}{180,0,0}      
\definecolor{mfk}{RGB}{0,100,0}      

\definecolor{noteyellow}{RGB}{255,186,0}	 
\definecolor{notegreen}{RGB}{0,128,0}        
\definecolor{noteblue}{RGB}{0,0,205}         
\definecolor{noteviolet}{RGB}{199,21,133}    

\usepackage{siunitx}
\usepackage{eurosym}
\usepackage{subcaption}
\usepackage{booktabs}
\usepackage[protrusion=true,expansion=true]{microtype}
\usepackage{hyperref}
\hypersetup{
  pdftitle={\ttitle},
  pdfauthor={Malte Kuhlmann, Ignacio Bugueno-Cordova, Emil Alms, Javier Ruiz-del-Solar, Nicolas Navarro-Guerrero},
  pdfsubject={Robotics (cs.RO); Artificial Intelligence (cs.AI)},%
  pdfkeywords={Pepper Robot, NVIDIA Jetson, Embedded Computation, ROS, D435i, Open Hardware, CAD, 3D Printing, robotic home assistant, Human-robot interaction, Social robotics, Speech recognition, LLMs, Chatbot, Object detection, Person detection, SLAM},%
  colorlinks=true,
  linktoc=all,
  linkcolor=darkgray,
  citecolor=darkgray,
  urlcolor=black,
}
\usepackage{paralist}

\usepackage{fp}
\FPeval{\totalsum}{round(400+60+346+150+60+1.3+1+5+5+3+15+2+5.5+12+26+30+3+1.4+0.8+0.2+0.2+0.2+0.2,0)}
\def\totalcost{$\num{\totalsum}$}

\theoremstyle{thmstyleone}%
\theoremstyle{thmstyletwo}%

\theoremstyle{thmstylethree}%

\begin{document}

\title{\textbf \ttitle}

\author[1]{\fnm{Malte} \sur{Kuhlmann}}\email{malte.kuhlmann@l3s.de}
\author[1]{\fnm{Ignacio} \sur{Bugueno-Cordova}}\email{i.bugueno@ieee.org}
\author[1]{\fnm{Emil} \sur{Alms}}\email{emil.alms@l3s.de}
\author[2,3]{\fnm{Javier} \sur{Ruiz-del-Solar}}\email{jruizd@ing.uchile.cl}
\author*[1]{\fnm{Nicolás} \sur{Navarro-Guerrero}}\email{nicolas.navarro.guerrero@gmail.com}

\affil*[1]{\orgname{Leibniz Universität Hannover}, \orgdiv{L3S Research Center}, \orgaddress{\street{Appelstraße 4}, \city{Hanover}, \postcode{30167}, \state{Lower Saxony}, \country{Germany}}}

\affil[2]{\orgname{University of Chile}, \orgdiv{Department of Electrical Engineering}, \orgaddress{\street{Avda.\ Tupper 2007}, \city{Santiago}, \postcode{8370451}, \state{Region Metropolitana}, \country{Chile}}}

\affil[3]{\orgname{University of Chile}, \orgdiv{Advanced Mining Technology Center}, \orgaddress{\street{Avda.\ Tupper 2007}, \city{Santiago}, \postcode{8370451}, \state{Region Metropolitana}, \country{Chile}}}

\abstract{
This paper presents the design, development, and experimental verification of {\addonname}, an open-hardware add-on for fast software development on the Pepper Robot. Our project enhances Pepper's computational and perception capabilities by integrating external components such as a Jetson Orin Nano, Logitech BRIO, Intel RealSense D435i, Samson UB1, and RØDE VideoMicro II. 
Our results show that the new hardware considerably improved Pepper's perception abilities and computational power. This development contributes to the community by implementing an open hardware and open-source modular add-on to the Pepper robot and keeping this relevant research platform functional beyond its expected lifespan. With {\addonname}, we aim to facilitate faster software development and more efficient integration of external components, ultimately enhancing the capabilities of the Pepper robot.
}


\maketitle

\section{Introduction}
\label{sec:introduction}
Several humanoid robot platforms have been extensively adopted in social robotics research. The NAO robot has featured in several research projects focused on education, assistive interaction, and long-term deployment worldwide \cite{Amirova2021NAOReview}. The iCub platform supports embodied cognition and manipulation research with open-source hardware and software \cite{Natale2017iCub}. Recent systems like Quori \cite{Specian2022Quori} and Reachy \cite{Sharma2025Robot} are purposely designed to offer modularity and affordability for HRI experimentation, combining expressive interaction with low cost.

\begin{figure}
    \centering
    \includegraphics[width=1\linewidth]{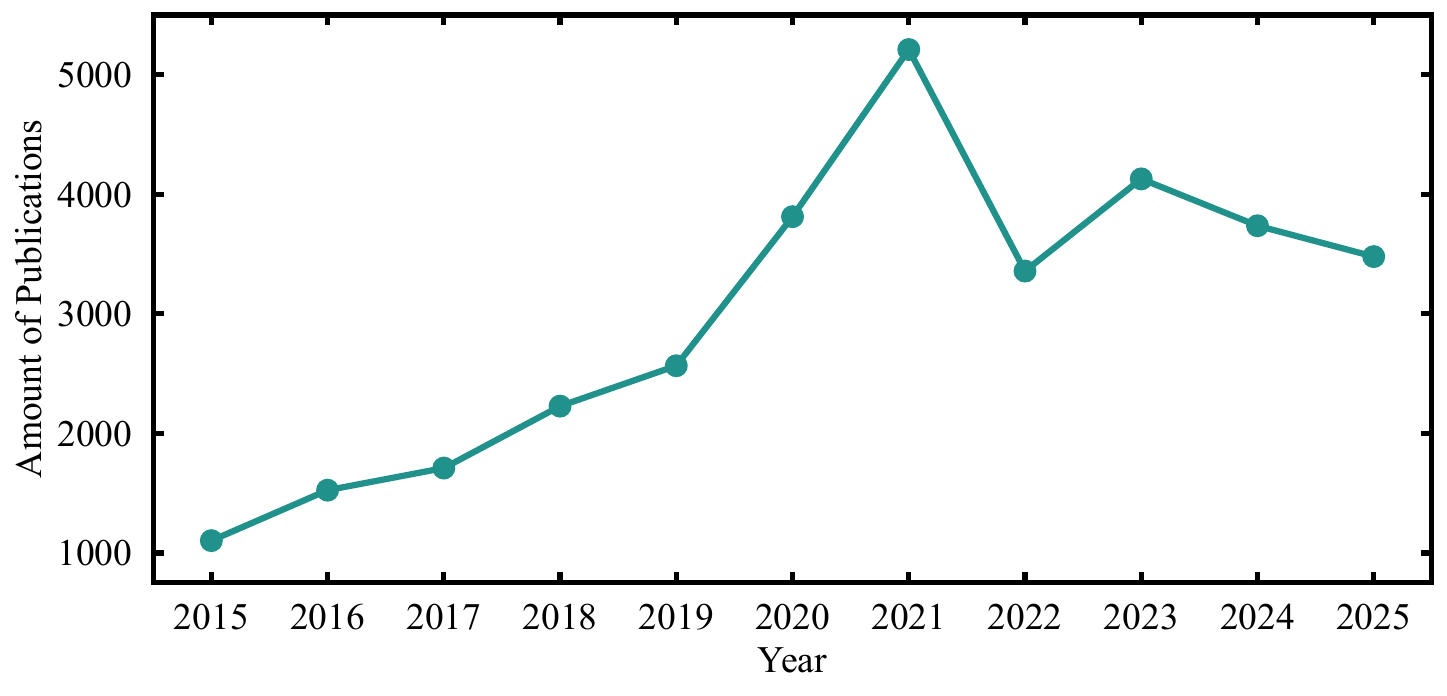}
    \caption{Number of publications mentioning the Pepper robot per year. We used the Dimensions dataset \cite{Adams2018Dimensions} to search for publications about the Pepper robot. Our search employed the query `Pepper AND Robot AND Social'.}
    \label{fig:yearly_research}
\end{figure}

Among these platforms, the Pepper robot has emerged as one of the most widely adopted humanoid systems for social robotics research and service applications~\cite{Pandey2018MassProduced}, as shown in Fig.~\ref{fig:yearly_research}. 
With over 27,000 units produced globally~\cite{nussey2021EXCLUSIVE}, Pepper has become a reference platform in HRI studies and real-world deployments in retail, hospitality, and education~\cite{Gardecki2018Pepper, Lier2019ToBI, DeCarolis2025Analyzing}. Recent research also show that Pepper remains relevant for modern embodied AI applications. Large Language Models have been explored for autonomous task execution and natural interaction~\cite{Rojas2026Improving}, while multimodal low-latency frameworks have been proposed for speech, perception, and robot control~\cite{Studerus2026Framework}.

However, its hardware and software pose significant limitations for advanced applications, particularly in terms of onboard computational power, restricted sensor quality, and limited perception capabilities~\cite{Reyes2019RealTime,Caniot2020Adapted}. Addressing these constraints expands Pepper’s applicability in complex and dynamic environments.

While cloud-based solutions that offload computationally intensive tasks to external systems are appealing, implementing such software on Pepper's onboard computer is becoming increasingly cumbersome as the platform ages. Also, some use cases, such as those related to navigation, require low latency to ensure safety, which cloud-based solutions cannot guarantee. Moreover, frequent connectivity issues can degrade Pepper's performance, especially when continuous data streaming is required. Additionally, data privacy and security concerns arise when external platforms process potentially sensitive audio or video streams, particularly in healthcare or educational environments.

Thus, other researchers have made hardware add-ons to overcome Pepper's computational and sensory limitations. Increasing the computation power of the robot also allows for integrating some cloud-based solutions. For instance, Caniot et al.~\cite{Caniot2020Adapted} integrated a Nvidia Jetson TX2 and an Intel RealSense D435i tested for thermal dissipation of the new hardware and robot's hardware, showing promising results. However, the authors did not release the mechanical or electrical design for replication.  

A similar concept introduced by Reyes et at.~\cite{Reyes2019RealTime} uses an Nvidia Jetson TK1 GPU mounted onto Pepper as a backpack. While this solution is open source, it only focuses on augmenting the onboard computational power of the robot but increases the robot footprint considerably.  

Bleth et al.~\cite{Bleth2023Improved} propose combining a LIDAR sensor with a Jetson Nano module, enabling Pepper to navigate more reliably in dynamic settings such as nursing homes. Increasing the resolution and range of Pepper's environmental perception mitigates the robot's risk of collisions and improves its capacity to move safely among people.

More recently, Magri et al.~\cite{Magri2025Upgrading} proposed an integrated hardware and software enhancement for Pepper by incorporating an NVIDIA Jetson GPU and an Intel RealSense D435i camera, enabling real-time perception algorithms such as human body orientation and gaze estimation. While this design improves the robot’s perceptual capabilities, it is limited in scope: it supports only a specific vision sensor and does not provide modular support for additional sensing modalities, such as external microphones. Moreover, its energy autonomy relies on an external battery, which can restrict continuous operation in mobile scenarios.

Previous studies have demonstrated the potential of open hardware in robotics \cite{Caniot2020Adapted, Mattamala2018NAO}. Moreover, robotics has seen significant advancements by integrating open hardware and software solutions, enabling faster development and more accessible innovation \cite{Patel2023Open}. Modular add-ons for existing robotic platforms are particularly interesting as they leverage the strengths of community-driven development to enhance robotic capabilities, in this case, a deprecated but a widely available robot~\cite{Pandey2018MassProduced}.

Hardware modifications such as integrating external GPUs or advanced sensors can considerably improve the robot's real-time perception and computational capabilities. Yet, these enhancements have challenges, including increased power consumption, thermal dissipation issues, and mechanical integration constraints. Ensuring that additional components do not disrupt Pepper's balance, mobility, or energy efficiency requires careful engineering. 

Inspired by these developments, we present {\addonname}, an open-hardware add-on powered directly by Pepper. It is specifically designed to support practitioners and developers in social robotics and HRI who continue to rely on Pepper as a standard platform but face growing limitations due to its ageing hardware. By enabling seamless integration of modern sensors and compute resources, {\addonname} helps bridge the gap between Pepper's original capabilities and the increasing algorithmic and perceptual demands of current research and applications. 
The open-source and hardware design (\href{https://creativecommons.org/licenses/by-nc-sa/4.0/deed.en}{CC BY-NC-SA 4.0}) is available at \url{https://github.com/mfkuhlmann/Pebre}.

The main contributions of this work are summarized as follows:
\begin{itemize}
    \item an open-hardware add-on powered directly by Pepper.
    \item a compact modular mechanical design based on custom 3D-printed components that preserves Pepper's mobility and form factor, while also allowing easy integration of different sensors on the head and around the tablet area.
    \item thorough technical validation covering thermals, wireless connectivity, audio, vision, and indoor mapping.
\end{itemize}

\section{{\addonname}: An Open-Hardware Add-On Design}
\label{sec:pepper-xp}
The design of {\addonname} was guided by the need to extend Pepper’s sensing and computational capabilities to meet the increasing demands of modern HRI and social robotics applications. In particular, we aimed to enable low-latency, safety-critical functionalities and responsive, interactive software pipelines that are not feasible with Pepper’s onboard hardware alone. A key requirement was to ensure seamless operation by powering the add-on directly from the robot, thereby avoiding additional power management complexity and reducing potential failure modes associated with maintaining multiple independent systems. At the same time, the platform was designed to support the flexible integration of heterogeneous sensors and computing modules, allowing developers to adapt the system to evolving research needs without repeatedly mounting and dismounting components, which could cause mechanical wear.

In addition to functional goals, several practical limits shaped the system design. Mechanically, the add-on must keep Pepper’s full movement and strength. In performance, it should not reduce the robot’s normal actions or disrupt its internal systems. The system should not raise Pepper’s internal temperature and have a low power consumption so as not to deteriorate Pepper’s autonomy. Finally, the design aims to have little effect on Pepper’s shape, so the add-on stays discreet and fits with the robot’s current build.

\subsection{Component Selection}
\label{sec:components}
The first step was to carefully select external hardware components to enhance Pepper’s functionality without compromising its existing system. Pepper’s CPU is already highly constrained, so adding new sensors and streaming their data quickly becomes taxing, limiting usable resolution and bandwidth. To address this, we integrated a dedicated computational unit inspired by prior work on extending Pepper. We selected the NVIDIA Jetson Orin Nano to process data from both new and onboard sensors while enabling communication with external computers and cloud services. The Jetson Orin Nano Super Developer Kit offers a good balance of compact size and computational power, while avoiding the need for a custom PCB and allowing installation behind Pepper’s tablet. Although a custom PCB could provide a more seamless integration, the developer kit significantly reduces development effort. In its standard 15 W power mode, the Jetson can be powered directly from Pepper without noticeably affecting autonomy, remaining well within the limits suggested by Caniot et al.~\cite{Caniot2020Adapted}. Power is supplied using a USB male-to-male Y-splitter combined with a DC/DC converter connected to the tablet.

Person and speech recognition are among the most common use cases for Pepper, yet its built-in microphones and cameras often underperform in real-world conditions. To address this, we selected lightweight, high-performance peripherals that can be integrated with minimal impact on the robot. Specifically, we designed and tested mounting points for the Samson UB1 and the RØDE VideoMicro II microphones for improved audio capture, as well as the Logitech BRIO and Intel RealSense D435i for enhanced visual perception. All sensors are directly connected to the Jetson, enabling efficient data processing.

We experimented with two microphones to improve audio capture: the Samson UB1 and the RØDE VideoMicro II. Table \ref{tab:microphone_comparison} provides a detailed comparison of these microphones. The selected microphones offer wider frequency ranges and higher sensitivity than the Pepper microphones. The choice of two different polar patterns allows for flexibility, catering to various use cases depending on the application's specific requirements.

\begin{table*}[tbp]
    \centering
    \setlength{\tabcolsep}{3pt}
    \renewcommand{\arraystretch}{1.2}
    \caption{Microphones Comparison.}
    \label{tab:microphone_comparison}
    \begin{tabular}{@{}cccc@{}}
        \toprule
         & Pepper & Samson UB1 & RØDE VideoMicro II \\ \midrule
         Frequency range & 100Hz to 10kHz & 30Hz - 18kHz & 20Hz - 20kHz  \\
         Sensitivity (dBV) & $-12$ & $-39$ & $-30$ \\
         Polar Pattern & Omni-Directional & Omni-Directional & Supercardioid \\
         \bottomrule
    \end{tabular}
\end{table*}

Similarly, the Logitech BRIO and Intel RealSense D435i cameras offer a significant improvement over Pepper's in-built cameras. The BRIO, mounted on Pepper’s head, is used for face and person recognition, while the RealSense D435i, mounted near the tablet, supports navigation and spatial perception. Table \ref{tab:camera_comparison_rgb} and Table~\ref{tab:camera_comparison_depth} present a feature comparison between these new and in-built cameras. Both upgraded cameras provide improved image quality and additional functionalities, enabling better performance for tasks requiring high-resolution or depth-sensing capabilities.

\begin{table}[htbp]
    \centering
    \setlength{\tabcolsep}{3pt}
    \renewcommand{\arraystretch}{1.2}
    \caption{Comparison between the RGB cameras.}
    \label{tab:camera_comparison_rgb}
    \begin{tabular}{@{}ccc@{}}
        \toprule
        & Pepper 2D & BRIO  \\ \midrule
        Mega Pixel & 5 & 13 \\
        FPS@1920*1080 & 15 & 60\\
        FOV (H x V) &  56.3° x 43.7° & 78° x 65° \\
        RGB Shutter & Rolling Shutter & Rolling Shutter  \\
        Focus Type & Auto Focus & Auto Focus \\
        \bottomrule
    \end{tabular}
\end{table}

\begin{table}[htbp]
    \centering
    \setlength{\tabcolsep}{3pt}
    \renewcommand{\arraystretch}{1.2}
    \caption{Comparison between the depth cameras.}
    \label{tab:camera_comparison_depth}
    \begin{tabular}{@{}ccc@{}}
        \toprule
        & Pepper 3D & D435i \\ \midrule
        Depth Resolution  & 1280 x 720 & 1280 x 720\\
        Depth FPS  & 15 & 90\\
        Depth FOV  & 96° x 60° & 87° x 58°\\
        Depth Shutter & Rolling Shutter & Global Shutter  \\ 
        Min.\ Distance (cm) & 40 & 28 \\
        \bottomrule
    \end{tabular}
\end{table}

\subsection{Mechanical Design and 3D Printing}
\label{sec:3D}
Custom mechanical parts were designed using computer-aided design (CAD) software to integrate into the Pepper robot. The designs were created with 3D printing in mind, resulting in low support waste, and are intended to use materials such as PLA for an easy and cheap print process. However, the head cover parts require PCTG, a stronger material, to ensure the durability of the hooks and parts on the head covers’ inner side. The process involved several iterations to overcome challenges related to the structural integrity of thinner components.

\begin{figure}[htbp]
    \centering
    \vspace{0.5em}
    \includegraphics[height=0.15\textheight]{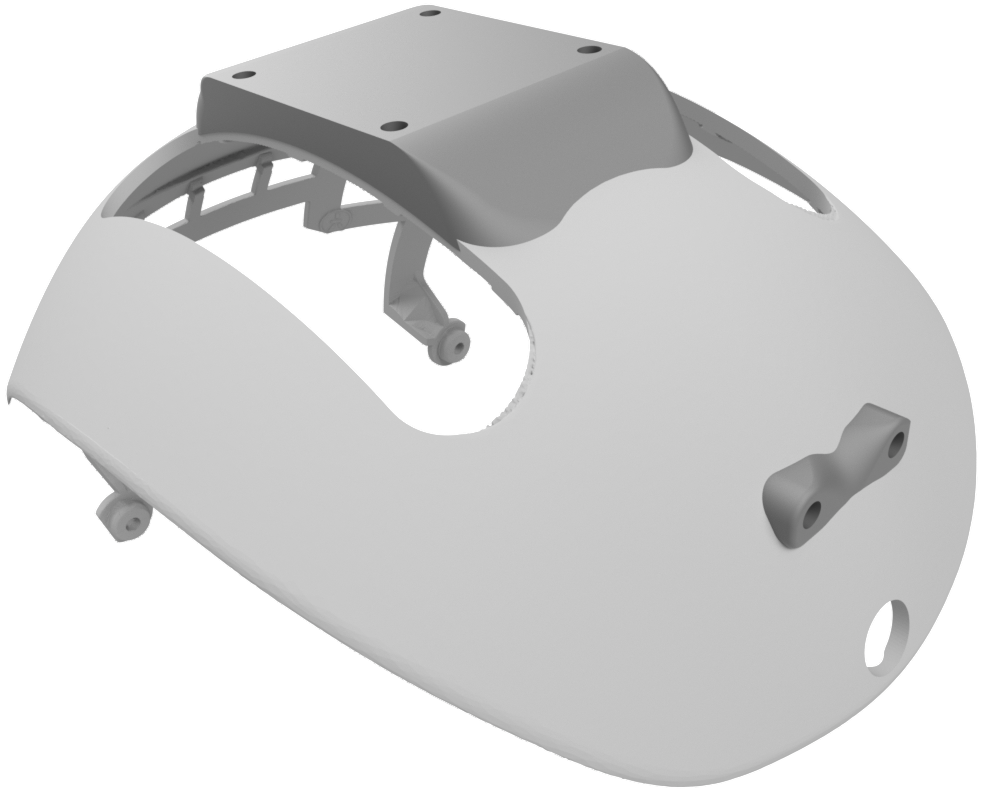}
    \includegraphics[height=0.15\textheight]{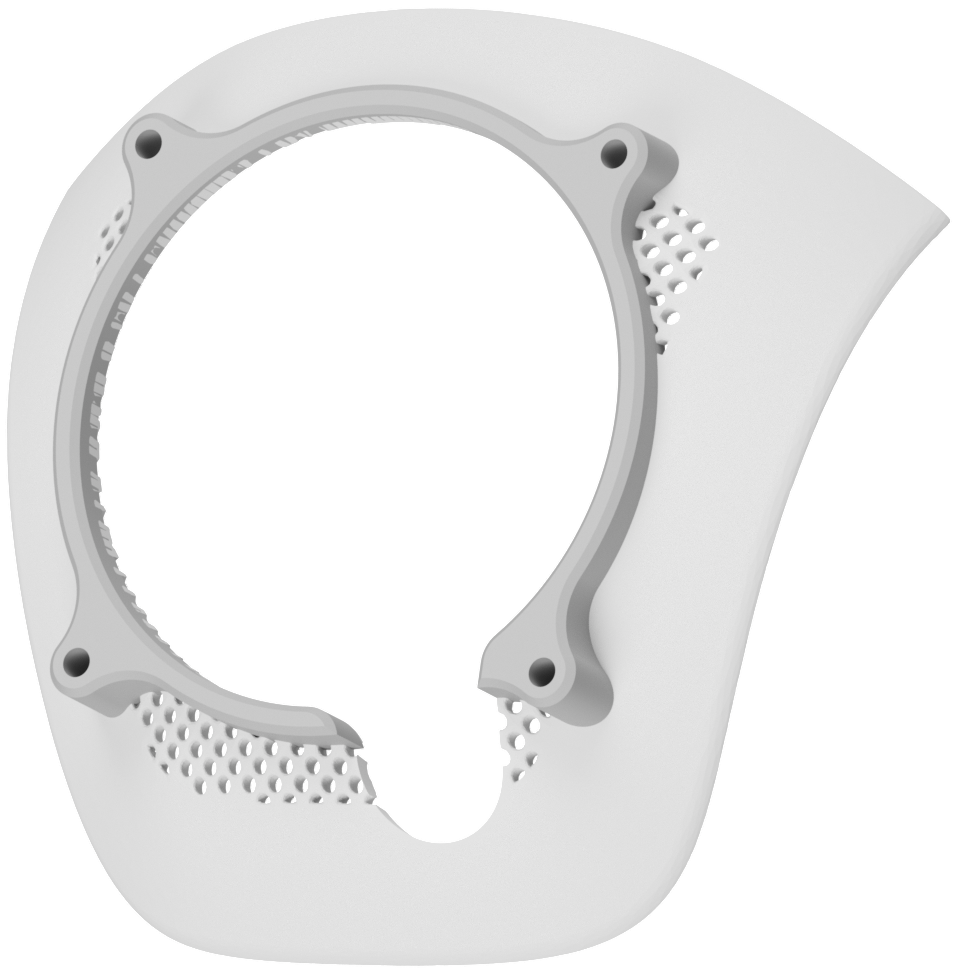}
    \includegraphics[height=0.15\textheight]{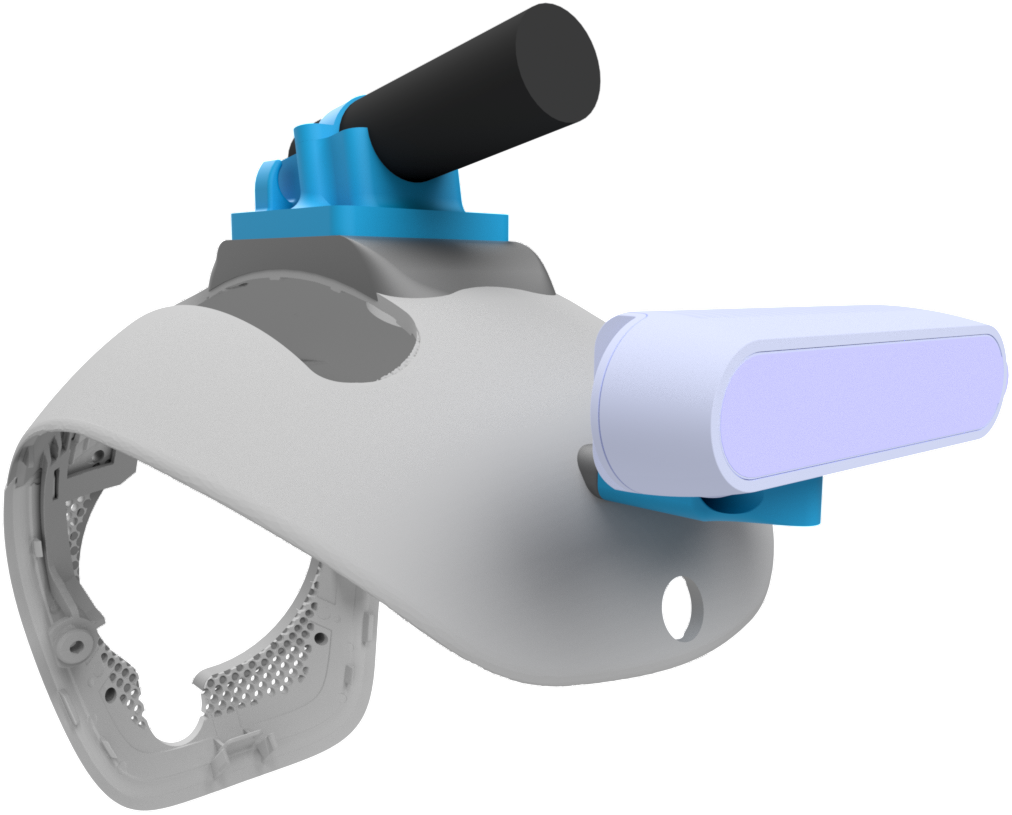}
    \includegraphics[height=0.15\textheight]{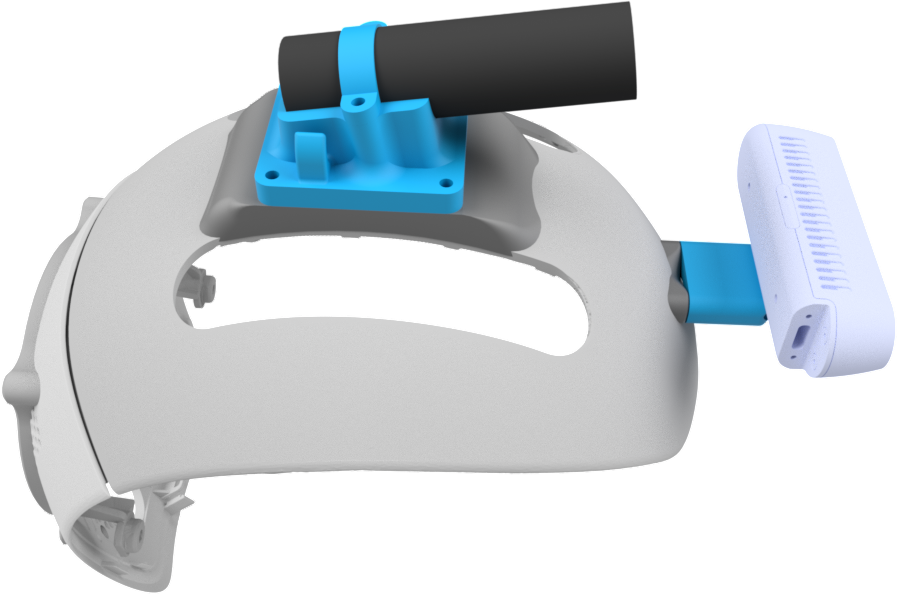}
    \caption{Customized head cover design with the RØDE VideoMicro II and the Intel RealSense D435i. Blue parts are removable.}
    \label{fig:head-covers}
\end{figure}

First, we designed new head covers to allow us to mount new hardware components and also improve air flow in Pepper's head. Due to the numerous small details, both head cover parts were digitally scanned. A Zeiss COMET 6 3D scanner, offering a resolution of 16 MPX and an accuracy of approximately $0.03mm$, was used for this purpose. The resulting design can be found in Fig.\ \ref{fig:head-covers}. 

\begin{figure}[htb]
    \centering
    \includegraphics[width=0.49\columnwidth]{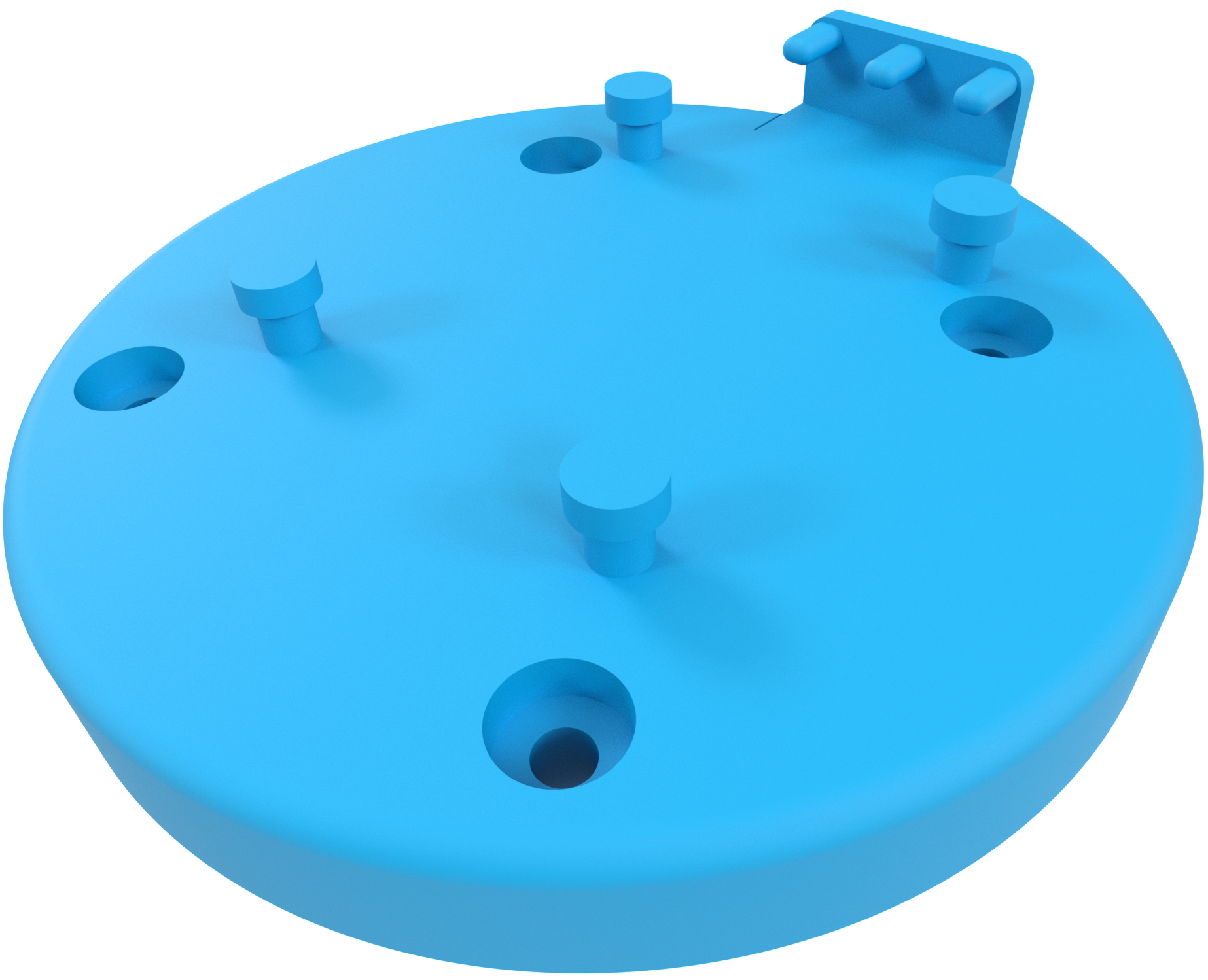}
    \includegraphics[width=0.49\columnwidth]{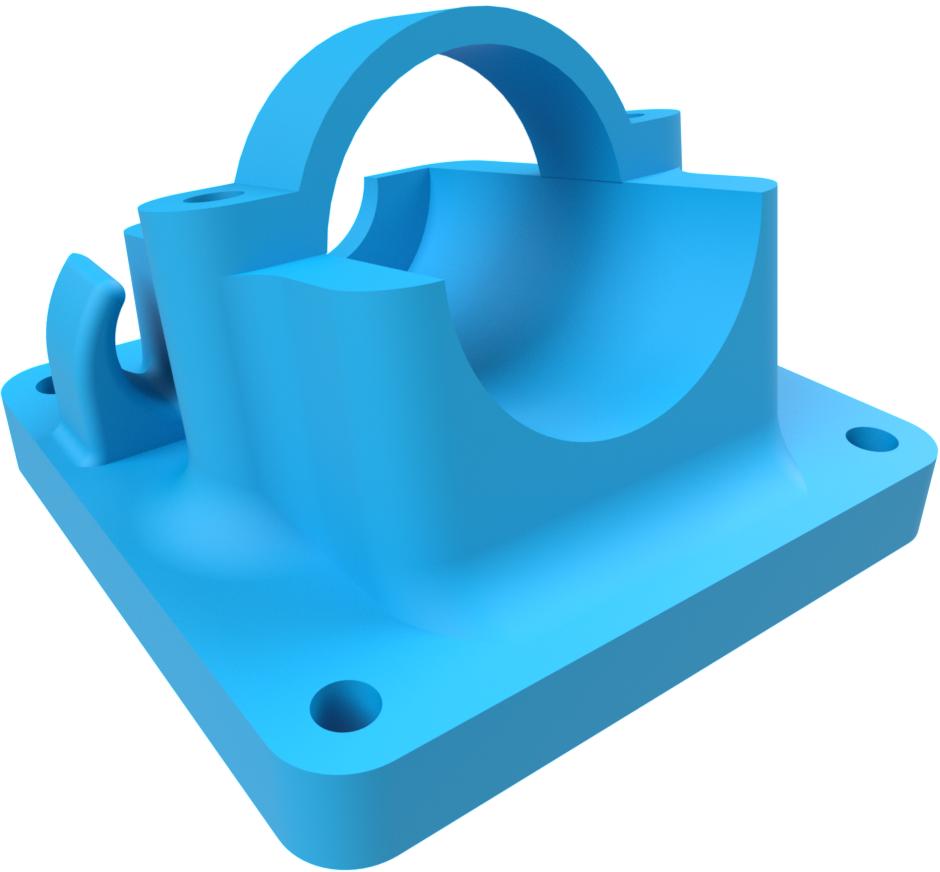}
    \includegraphics[height=0.15\textheight]{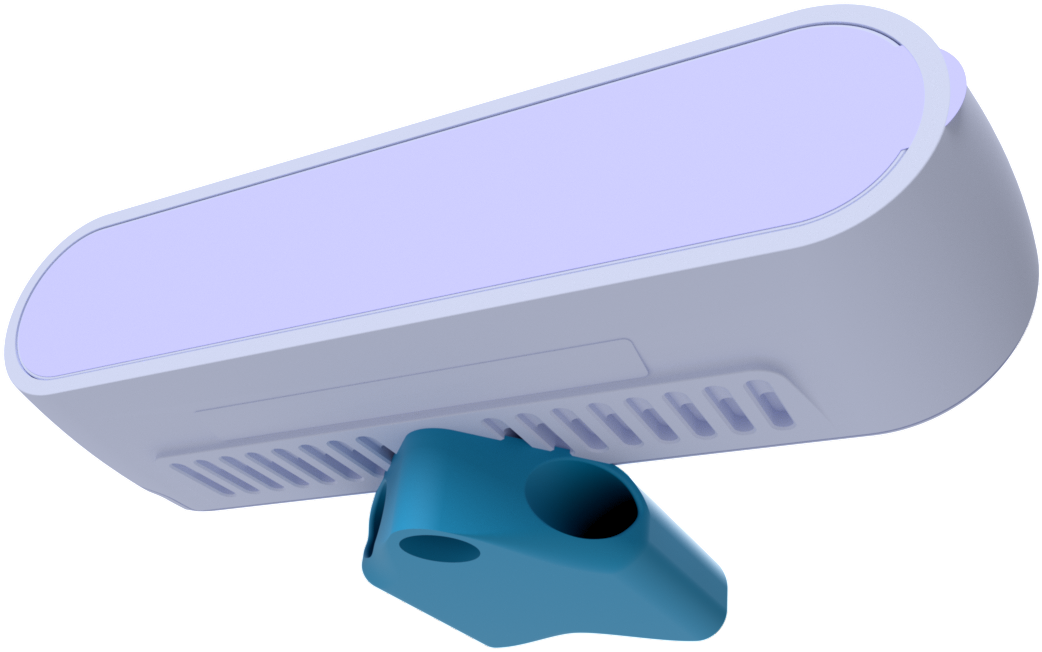}
    \caption{Interchangeable Microphone Holders: Samson UB1 (Left) and RØDE VideoMicro II (Right). Camera holder capable of supporting either a Logitech BRIO or an Intel RealSense D435i.}
    \label{fig:head-holder}
\end{figure}

We modified the top head cover to place the microphone and camera, which can be removed without unmounting the head covers. The necessary holders are shown in Fig.\ \ref{fig:head-holder}. We tested two microphones. The Samson UB1 was mounted on top of the head to keep a low center of mass, while the RØDE VideoMicro II was pointed forward for improved speech recognition in noise environments. The Logitech BRIO camera is placed above Pepper's in-built camera, aligning its view with Pepper's native perspective.

\begin{figure}[htbp]
    \centering
    \includegraphics[width=0.49\columnwidth]{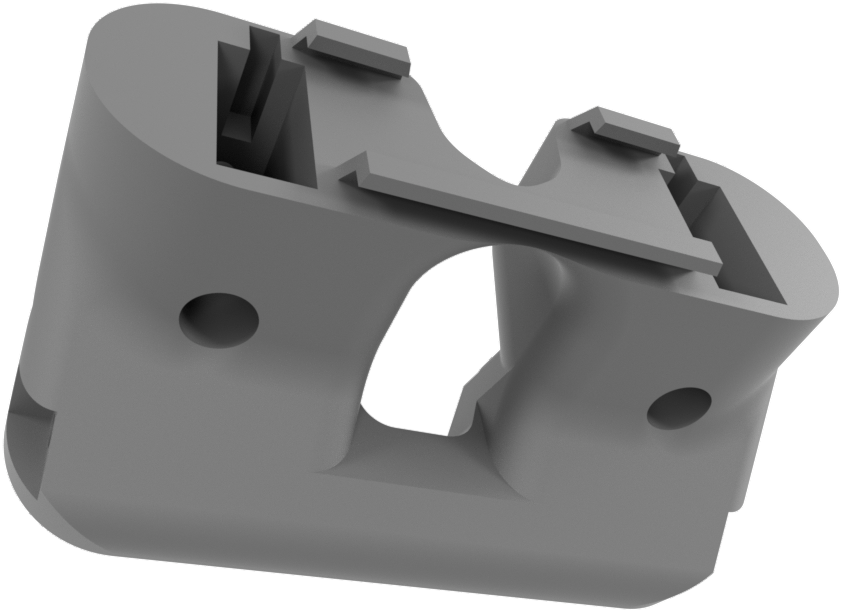}
    \includegraphics[width=0.49\columnwidth]{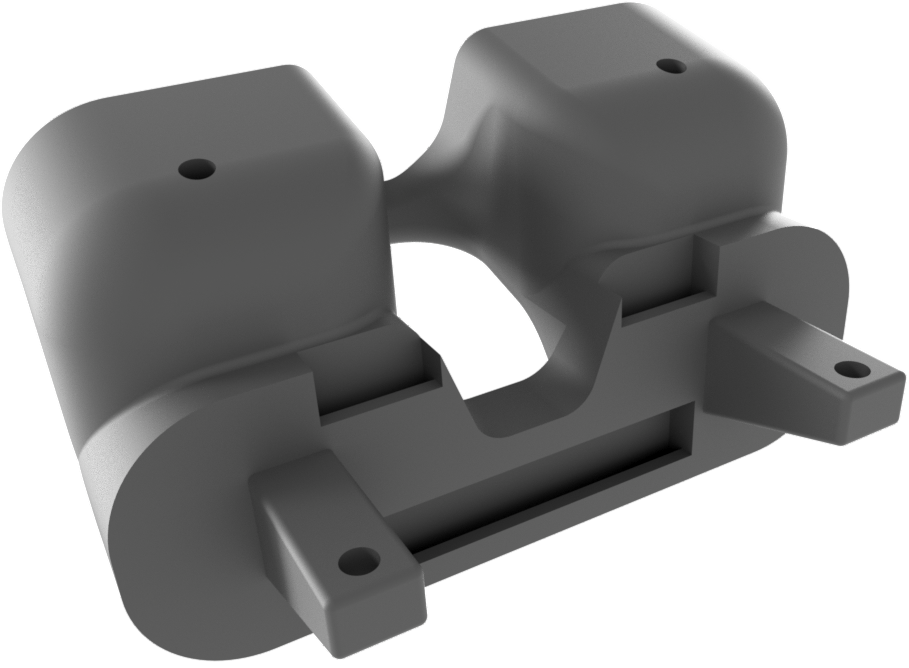}
    \caption{Tablet holder design featuring mounting points for both the Jetson enclosure and the Pepper tablet. The central opening enables power bifurcation via a USB cable.}
    \label{fig:tablet-holder}
\end{figure}

In addition, we designed a \textit{Tablet Holder}, which can be mounted between the robot's chassis and Pepper's tablet. It allows the \textit{Jetson Box} to be mounted to Pepper. Additionally, turning the tablet 15 Degrees to improve viewing angles. Fig.~\ref{fig:tablet-holder} contains a close-up of this \textit{Tablet Holder}. We ensured that the new tablet's position did not cause the head or arms to collide with the tablet.

\begin{figure}[htbp]
    \centering
    \vspace{0.5em}
    
    \begin{minipage}{\linewidth}
        \centering
        \includegraphics[height=0.2\textheight]{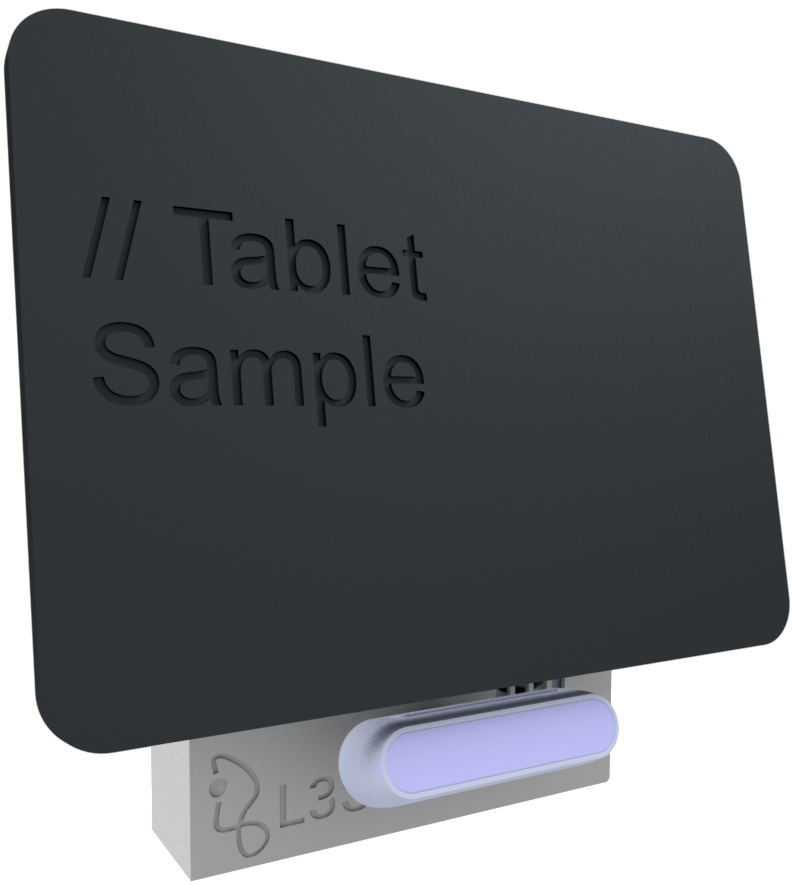}
        \vspace{0.5em}
    \end{minipage}
    
    \begin{minipage}{0.45\linewidth}
        \centering
        \includegraphics[height=0.2\textheight]{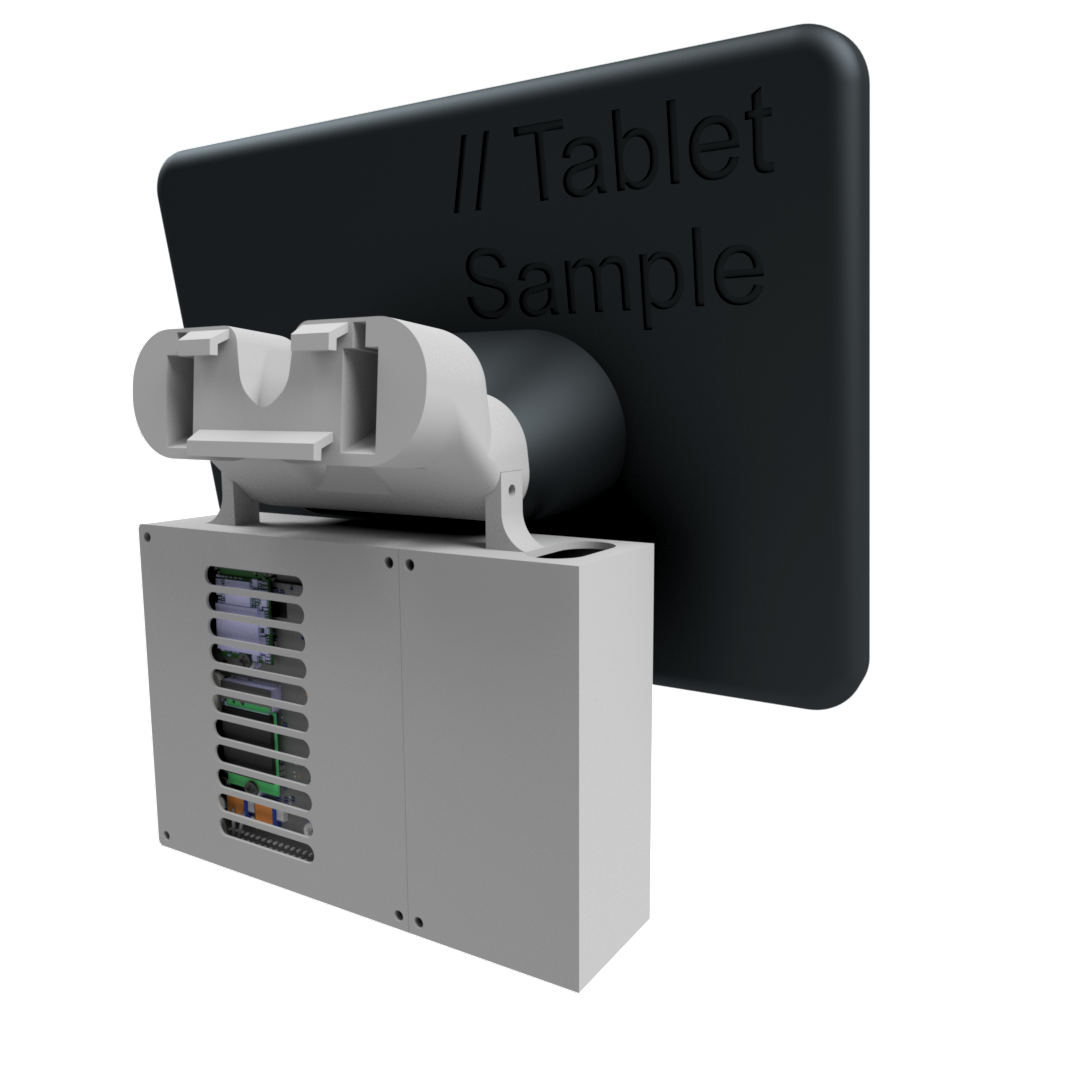}
    \end{minipage}
    \hfill
    \begin{minipage}{0.45\linewidth}
        \centering
        \includegraphics[height=0.2\textheight]{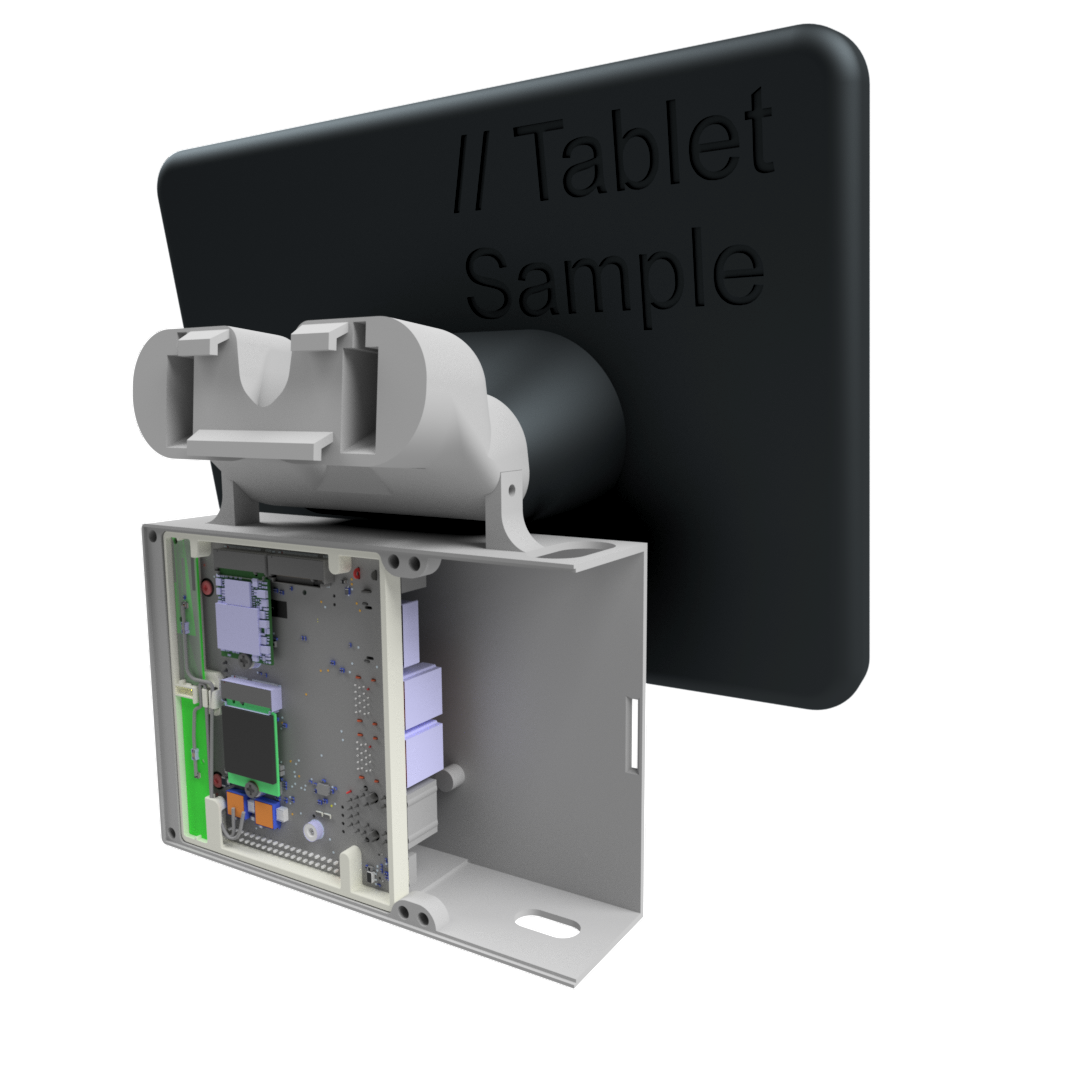}
    \end{minipage}
    \caption{\textit{Jetson Box} design.}
    \label{fig:jetson-box}
\end{figure}

\begin{figure}[htbp]
    \centering
    \includegraphics[width=0.5\linewidth]{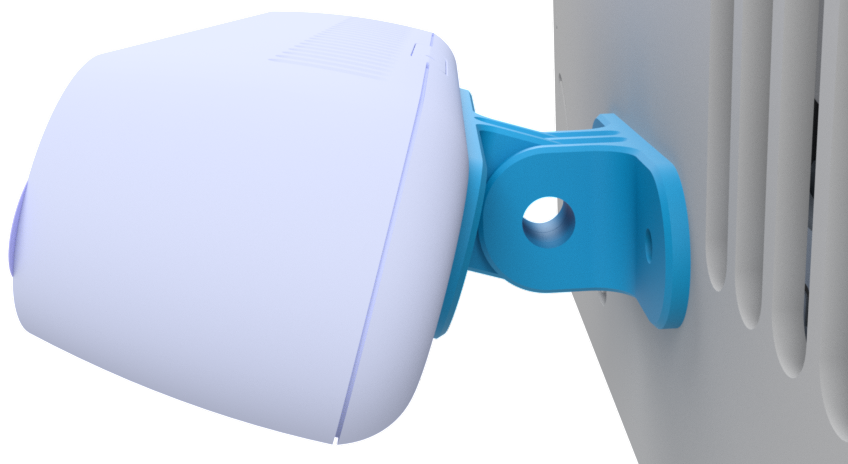}
    \caption{Tilting mechanism for the Intel RealSense D435i.}
    \label{fig:angular_lock}
\end{figure}

\begin{figure}[htbp]
    \centering
    \includegraphics[width=0.5\columnwidth]{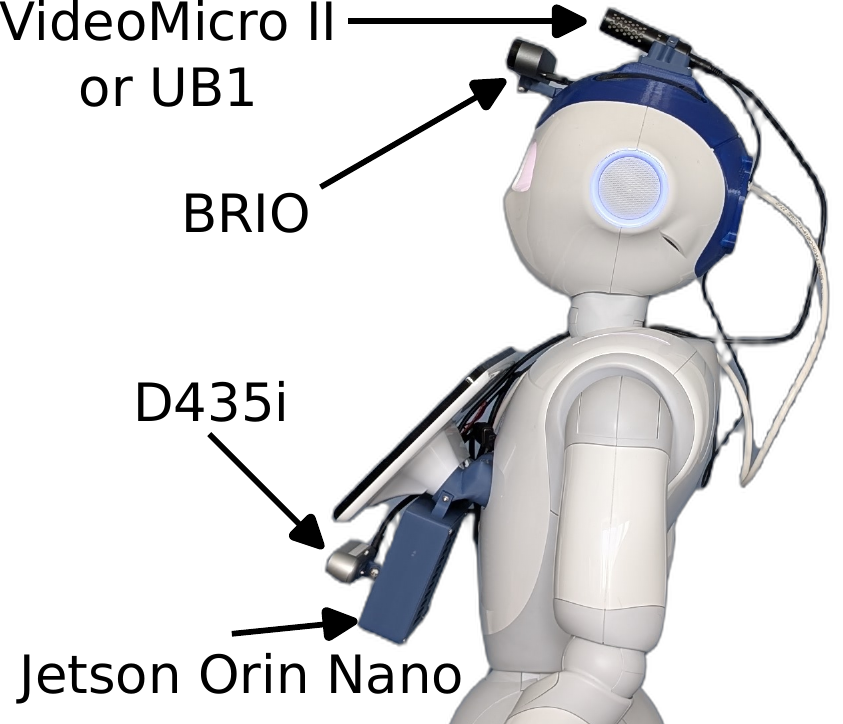}
    \caption{Assembled {\addonname} showing the location of the selected components.}
    \label{fig:PEBRE}
\end{figure}

Complementary to the \textit{Tablet Holder}, we designed a box that holds the Jetson Orin Nano. The Jetson Orin Nano is positioned in the box behind the tablet to ensure it remains partially hidden and to keep the robot's footprint almost unaltered. Furthermore, the box provides a compartment to store excess cables and fully encloses the Jetson to protect it from unwanted impact, while still delivering optimal cooling performance through air inlets. This placement is shown in Fig.~\ref{fig:jetson-box}. The Intel RealSense D435i is mounted on the \textit{Jetson Box} through a tilting mechanism, allowing for a precise position of the camera for various scenarios and providing an unrestricted view of the environment. The design for the pivot can be seen in Fig.~\ref{fig:angular_lock}. Fig.~\ref{fig:PEBRE} shows the assembled add-on indicating the location of the selected components.

\subsection{Software Environment Setup}
\label{sec:software}
A crucial aspect of this project was setting up a stable software environment that could support the integration of the selected hardware components. This involved configuring the necessary drivers, libraries, and frameworks to ensure compatibility with the Pepper robot's existing system. The goal was to create a seamless interaction between the external components and the robot's software, facilitating future developments and applications.

Python was selected as the programming language for this project due to its intuitive nature and robust support for artificial intelligence applications and ROS(2). It is well-suited for handling complex interactions between hardware components and software frameworks. A Python 3.10 environment was created on the Jetson Orin Nano. This choice aimed to balance functionality by supporting legacy and modern Python packages. While more recent Python versions can enhance operational speed for some applications, older versions, until Python 3.7, can also be used in case of backward compatibility~issues.

The environment's initial setup involved installing the Qi framework \cite{libqi} and the Qi Python wrapper. These tools enable communication between the Jetson Orin Nano and the Pepper robot, allowing access to the robot's services through Python~3. The package was compiled from source code into a wheel format, ensuring easy installation via PIP.

Subsequently, supplementary hardware components were configured. The Samson UB1, RØDE VideoMicro II and Logitech BRIO integrated seamlessly with the Jetson Orin Nano without requiring additional drivers. However, the Intel RealSense D435i camera presented a challenge: it needed the realsense2 library \cite{librealsense} and the pyrealsense2 Python wrapper to operate effectively, which are currently not supported on Jetpack Version 6.0 or higher. This issue was successfully resolved by compiling the library with a legacy backend.

\section{Validation Methodology}
\label{sec:methodology}
To evaluate the contributions of {\addonname} add-on, we perform a comprehensive validation framework structured as follows: mechanical and hardware validation, wireless connectivity, sensor comparison, and functional capabilities. All experiments are designed to compare the enhanced Pepper configuration against the original robot baseline, highlighting both quantitative improvements and new functionalities enabled by the proposed add-on.

\subsection{Mechanical and Hardware Validation}
From a mechanical perspective, the custom-printed head covers, holders, tablet extension, and Jetson enclosure are inspected for structural integrity under typical handling and robot motions.  
Their integration is verified to avoid collisions or restrictions of the robot’s joints and to ensure that the additional payload does not compromise Pepper's kinematics or stability.

\subsection{Thermal Validation}
The hardware assessment focuses on both the Jetson~Orin~Nano and Pepper's internal processor, as their performance can be constrained by heat accumulation within their respective enclosures.  

For the Jetson~Orin~Nano, two deployment configurations are compared:
\begin{inparaenum}[a)]
  \item operation in open air on a desk, and
  \item operation inside the 3D-printed enclosure attached to Pepper.
\end{inparaenum}
A sustained stress workload is generated using the YOLOv8n network in inference-only mode at an input resolution of \(2048\times2048\) pixels with a batch size of~8.  
The test is executed in two variants CPU-bound and CUDA-accelerated, thereby stressing both principal processing subsystems.  
Temperature measurements are collected continuously during inference under both configurations, allowing direct comparison of the thermal impact introduced by the enclosure and its airflow characteristics.

A similar stress test is performed on Pepper’s internal processor (Intel~Atom~E3845) to evaluate the thermal stability of the robot’s head.  
The test applies a sustained 100\% CPU workload and samples per-core thermal data at 1Hz.  
Three configurations are examined:
\begin{inparaenum}[a)]
  \item the head without its enclosure (open configuration),
  \item the head with the original plastic enclosure, and
  \item the head with the {\addonname} enclosure.
\end{inparaenum}
These measurements enable a quantitative assessment of how the add-on design influences heat dissipation and internal airflow, complementing the thermal validation performed on the Jetson~Orin~Nano.

\subsection{Wireless Interfaces}
Wireless connectivity affects both remote robot teleoperation and the transfer of large sensor data streams. 
We evaluate the Wi-Fi interface of the Jetson Orin Nano installed in {\addonname} in comparison with Pepper’s internal wireless adapter.

The robot is positioned at fixed distances $d \in \{1, 2, 4, 8, 16, 32\}\,\mathrm{m}$ from a Wi-Fi access point under two environmental conditions:
\begin{inparaenum}[a)]
  \item \emph{line-of-sight} (LoS), in an open corridor without physical obstructions, and
  \item \emph{obstructed} (Obs), where the router is placed just behind the closed door of a 10\,m\,$\times$\,10\,m room with 20\,cm-thick interior walls.
        The Pepper and Jetson Orin Nano are positioned outside the room at the specified distances, so that the signal must pass through the door and wall, creating a controlled, strong-attenuation scenario.
\end{inparaenum}

To emulate representative data traffic, four dummy files of increasing size are transmitted from each platform to a LAN-connected notebook using \texttt{scp}: 
a command file (\texttt{.txt}, 0.01\,MB), an audio clip (\texttt{.wav}, 2\,MB), an image (\texttt{.png}, 5\,MB), and a LIDAR-like binary log (\texttt{.bin}, 50\,MB).

Measurements of network \emph{ping} latency were discarded, as they exhibited high short-term variability and showed no systematic variation across the tested distances. 
Even the largest separation in our setup (32 m) is too short to capture distance-dependent propagation effects in typical indoor Wi-Fi channels, making latency unsuitable for meaningful channel-analysis in this context.
Instead, two metrics are analyzed:
\begin{inparaenum}[a)]
  \item the received signal strength indicator (RSSI) at the platform’s wireless interface, and
  \item the file-transfer time $T_{f}$ for each file type.
\end{inparaenum}

For each distance and condition, RSSI is sampled at 1 Hz for a duration of $T_\mathrm{RSSI} = 30\,\mathrm{s}$. 
Each sample represents the logarithmic received signal power
\begin{equation}
    R(t) = 10 \log_{10} \!\left(\frac{P_\mathrm{rx}(t)}{1\,\mathrm{mW}}\right) \, [\mathrm{dBm}],   
\end{equation}
where $P_\mathrm{rx}(t)$ is the instantaneous received power at time $t$.  
The samples over each run are summarized by their mean and standard deviation to characterize the average channel strength and its short-term fluctuations.

File-transfer performance is assessed by repeating each file transmission $K = 30$ times.  
For each repetition, the transfer time is defined as $T_{f}^{(k)} = t_{\mathrm{end}}^{(k)} - t_{\mathrm{start}}^{(k)},$
where $t_{\mathrm{start}}^{(k)}$ and $t_{\mathrm{end}}^{(k)}$ are the timestamps of transmission start and completion.  
The resulting set of $K$ measurements is also summarized by its mean and standard deviation.

\subsection{Microphones}
To evaluate sensitivity and directional response, three microphones are analyzed: the built-in unit on Pepper, the Samson UB1 (omnidirectional), and the RØDE VideoMicro II (directional). 
A broadband frequency sweep ranging from 20\,Hz to 10\,kHz is emitted by a loudspeaker positioned 1.0\,m in front of the microphones. 
Recordings are acquired simultaneously for all three microphones at angular offsets of 0\textdegree, 45\textdegree, and 90\textdegree relative to the sound source. 

The first five seconds of each recording, before sweep onset, are used to estimate baseline noise. 
Each signal $x[n]$ is transformed into the frequency domain using a Hann window $w[n]$ and the single-sided FFT: 
\begin{equation}
X[k] = \frac{1}{N \, G} \sum_{n=0}^{N-1} x[n] w[n] e^{-j 2 \pi kn/N},
\end{equation}
where $G = \tfrac{1}{N}\sum_{n=0}^{N-1} w[n]$ is the coherent gain of the window.  
The corresponding amplitude spectrum in dBFS is obtained as $M[k] = 20 \log_{10} \big( |X[k]| \big).$
These spectra are plotted across the range 0–10\,kHz to visualize the relative frequency response of each microphone at different orientations.  

For noise–signal characterization, Welch’s method is used to estimate power spectral density $P_{xx}(f)$ within a given frequency band $\mathcal{B}$: 
\begin{equation}
P_{\mathcal{B}} = \int_{\mathcal{B}} P_{xx}(f) \, df.
\end{equation}
Noise power $P_{N}$ is computed from the initial baseline segment, and signal power $P_{S}$ from subsequent windows.  
The signal-to-noise ratio (SNR) in dB is defined as
\begin{equation}
\mathrm{SNR}_{\mathcal{B}} = 10 \log_{10} \left( \frac{P_{S}}{P_{N}} \right),
\end{equation}
evaluated specifically for the speech-relevant band $\mathcal{B} = [300, 3400]$\,Hz.  

For each microphone and orientation, SNR is evaluated over overlapping temporal windows of 1.0\,s with a 0.5\,s hop size to capture local fluctuations during the sweep.  
The resulting SNR values across all windows are summarized by their mean and standard deviation, providing an overall measure of average signal quality and its variability for each microphone–orientation pair.

\subsection{Cameras}
Perceptual improvements introduced by {\addonname} are evaluated by comparing Pepper's built-in RGB–D sensor against the integrated Logitech BRIO and Intel RealSense~D435i.
\subsubsection{RGB Cameras}
To assess the relative image quality of Pepper's built-in RGB sensor and the Logitech BRIO, a face recognition pipeline was implemented and evaluated under controlled conditions.
\paragraph{Data Acquisition}
A database of 3 facial identities serves as the recognition reference. To evaluate system robustness, six acquisition scenarios were defined, each capturing subjects already present in the database under distinct perceptual conditions:

\begin{itemize}
\item \textbf{Frontal}: The subject faces the robot directly; both parties remain stationary.
\item \textbf{Human head movement}: The subject stands in close proximity to the robot and rotates their head laterally at a moderate pace while the robot remains stationary.
\item \textbf{Pepper head movement}: The inverse of the previous scenario. The subject remains stationary while Pepper's head rotates laterally at a moderate pace.
\item \textbf{Distance}: The subject stands approximately 4.5~m from the robot; neither party moves.
\item \textbf{Overexposure}: The subject is positioned directly in front of a window, producing strong backlighting and significant overexposure.
\item \textbf{Low lighting}: Windows are occluded and ambient illumination is reduced to a minimum.
\end{itemize}
Both cameras capture data simultaneously to ensure identical environmental conditions across comparisons. For each scenario, a 10-second video is recorded and uniformly subsampled to 10 frames. This results in $180$ total frames.

\paragraph{Face Recognition}
The sampled frames are processed by the face detection and recognition models provided by the \texttt{dlib} library~\cite{dlib09}, specifically the hog-based face detector and the compact face encoding model. Each frame is first passed through the detector. Upon a successful detection, the cropped face region is forwarded to the encoder to produce a feature embedding. The cosine similarity between this embedding and a pre-encoded reference image is then computed, and a match is declared if the similarity exceeds a threshold of $0.6$. Camera quality is assessed in terms of face detection rate and face recognition accuracy across scenarios.

\subsubsection{Depth Cameras}
Since Pepper's proprietary software lacks native dense-mapping support, a dedicated data acquisition and 3D reconstruction pipeline was developed for both sensor configurations to ensure a controlled comparison.
\paragraph{Data Acquisition}
During baseline experiments, Pepper is teleoperated along a fixed indoor trajectory while recording:
\begin{inparaenum}[a)]
stereo camera images and
IMU measurements. 
\end{inparaenum}
Data is stored in TUM dataset format. We collected the recording at Pepper's maximum motion speed. Each consists of multiple loops along the predefined path. A fixed head pose is maintained throughout to eliminate extrinsic variation between runs.
\paragraph{Mapping}
Dense maps are constructed offline using ORB-SLAM3 \cite{Campos2021ORBSLAM3}. Camera calibration is performed prior to mapping: the Intel RealSense D435i provides factory intrinsics automatically, whereas Pepper's built-in camera is calibrated manually using Kalibr with an AprilTag grid pattern ($21$~mm tag size). IMU-to-camera extrinsic parameters are also estimated using Kalibr with the same AprilTag grid pattern.

\subsubsection{Comparative evaluation}
The mapping task itself serves as a representative benchmark to expose differences in sensor quality, since it integrates spatial, temporal, and photometric aspects of perception into a downstream robotics application.  
The two depth sensors are contrasted in terms of hardware characteristics, see Table~\ref{tab:camera_comparison_depth}, including resolution, maximum frame rate, field of view, shutter type, and minimum distance.

To complement the geometric reconstruction results, per-frame image quality metrics are computed for both RGB and depth data. A small static dataset is captured for this. All metrics are calculated for each frame and then averaged over the full sequence.

\subparagraph{Metric definitions for RGB frames}
The mean luminance $\mu_{Y}$ and the luminance standard deviation $\sigma_{Y}$ are computed on the ITU-R~BT.601 gray-level channel
$
    Y = 0.299\,R + 0.587\,G + 0.114\,B,
$
scaled to the $[0,255]$ range.

The luminance entropy $H_{Y}$ is defined as the Shannon entropy of the normalized gray-level histogram $H_{Y} = - \sum_{i} p_{i} \log_{2} p_{i}$,
where $p_{i}$ denotes the empirical probability of bin $i$.

The under-exposed and over-exposed pixel fractions are computed as the proportions of pixels with $Y \leq 5$ and $Y \geq 250$, respectively.

The sharpness is quantified as the variance of the Laplacian operator applied to the normalized luminance image $\mathcal{S} = \mathrm{Var}\bigl( \nabla^{2} Y \bigr)$.

The colorfulness is estimated by the Hasler--Süsstrunk metric:
\begin{equation}
    \mathcal{C} = \sqrt{\sigma_{rg}^{2} + \sigma_{yb}^{2}}
    + 0.3\sqrt{\mu_{rg}^{2} + \mu_{yb}^{2}},
\end{equation}
where $rg = R - G$ and $yb = 0.5(R + G) - B$.

\subparagraph{Metric definitions for depth frames}
The mean $\mu_{D}$ and standard deviation $\sigma_{D}$ of valid raw depth values are reported. The depth entropy $H_{D}$ is defined analogously to $H_{Y}$ over the depth histogram.

The valid pixel ratio $r_{\mathrm{valid}}$ and invalid pixel ratio $r_{\mathrm{invalid}}$ are computed as the proportions of pixels with non-zero and zero depth, respectively.

The low- and high-saturation fractions are defined as the proportions of pixels falling below a fixed low-depth threshold or exceeding a near-maximum representable depth value.

The gradient magnitude mean is obtained by computing the Sobel gradient $\nabla D$ on a normalized depth map and averaging its magnitude:
\begin{equation}
    \bar{G} = \frac{1}{N}\sum_{u,v} \|\nabla D(u,v)\|,
\end{equation}
which captures local surface variations. 
Finally, the bit depth indicates the numeric precision of stored pixel values. 
These metrics collectively characterize photometric quality for RGB sensing and geometric fidelity for depth sensing under typical indoor illumination conditions.

\section{Results}
\label{sec:results}
This section reports the results obtained after the systematic application of the methodology described in Section~\ref{sec:methodology}.  
The results are organized into
\begin{inparaenum}[(i)]
    \item deployment of {\addonname},
    \item mechanical and hardware validation,
    \item thermal validation, 
    \item wireless performance, 
    \item microphones,
    \item RGB and depth cameras. 
\end{inparaenum}

\subsection{Deployment}
{\addonname} was fabricated and assembled for integration on the Pepper robot.  
Structural parts were produced on a Prusa CORE One 3D printer, requiring approximately $210$\,g of polylactic acid (PLA) filament printed with around 20\% infill, utilizing a 0.4mm Nozzle. Specialized parts, used for the head covers, were printed in Polycyclohexylendimethylenterephthalat-Glykol (PCTG) to improve layer adhesion, weighing in at around $160$\,g of Filament. In addition to the different filament, the nozzle was also changed to a 0.25mm diameter, to provide more detail and better strength, however a 0.4mm nozzle should also be sufficient. For a detailed line up of materials used and support waste see Table~\ref{tab:parts}.  
Commercial components, including computing units, sensors, audio devices, fasteners, and cabling, were procured for assembly at a total cost of about \totalcost \euro{}, as listed in Table~\ref{tab:Bill of materials}.  
Together, these tables summarize the material requirements for reproducing the add-on.

The completed assembly is illustrated in Fig.~\ref{fig:motion_clearance}, showing the robot equipped with the mounted Jetson enclosure, and the front-mounted microphone and vision sensors. 

\begin{table}[htbp]
    \footnotesize
    \centering
    \setlength{\tabcolsep}{3pt}
    \renewcommand{\arraystretch}{1.2}
    \caption{3D Printed parts and quantity of filament.}
    \label{tab:parts}
    \begin{tabular}{@{}c|c|c|c@{}}
       \toprule
       \textbf{No.} & \textbf{Component} & \textbf{Part (g)} & \textbf{Waste (g)} \\ \midrule
       \multicolumn{4}{c}{\textbf{PLA}} \\ \midrule
        1 & Jetson Box\phantom{*} & 80.21 & 3.40 \\ \hline
        2 & Jetson Back Cover\phantom{*} & 13.78 & 0.00 \\ \hline
        3 & Jetson Cable Cover\phantom{*} & 9.38 & 0.00 \\ \hline
        4 & Pepper Tablet base extension\phantom{*} & 39.33 & 10.33 \\ \hline
        5 & Intel RealSense Holder P1\phantom{*} & 2.78 & 0.00 \\ \hline
        6 & Intel RealSense Holder P2\phantom{*} & 2.70 & 0.00 \\ \hline
        7 & Samson UB1 Microphone P1 & 16.39 & 0.00 \\ \hline
        8 & Samson UB1 Microphone P2 & 1.27 & 0.00 \\ \hline
        9 & RØDE VideoMicro II P1 & 10.72 & 0.00 \\ \hline
        10 & RØDE VideoMicro II P2 & 19.38 & 0.00 \\ \hline
        11 & Top Camera Holder & 3.33 & 0.00 \\ \hline
        12 & Cable Holder Neck & 2.10 & 0.00 \\ \midrule
        \multicolumn{2}{l|}{\textbf{Total}} & $\sim192.98$ & $\sim13.73$ \\ \midrule
        \multicolumn{4}{c}{\textbf{PCTG}} \\ \midrule
        13 & Pepper Head Rear Cover & 32.57 & 14.48 \\ \hline
        14 & Pepper Head Front Cover & 82.87 & 29.18 \\ \midrule
        \multicolumn{2}{l|}{\textbf{Total}} & $\sim115.44$ & $\sim43.66$ \\ \bottomrule
    \end{tabular}
\end{table}

\begin{table}[htbp]
    \footnotesize
    \centering
    \setlength{\tabcolsep}{3pt}
    \renewcommand{\arraystretch}{1.2}
    \caption{Bill of materials.}
    \label{tab:Bill of materials}
    \begin{tabular}{@{}c|c|c|c@{}}
        \toprule
        \textbf{No.} & \textbf{Component} & \textbf{Units} & \textbf{Price (\euro{})} \\
        \midrule
        1 & NVIDIA Jetson Orin Nano & 1 & $\sim400$ \\ \hline
        2 & SSD PCIe NVMe & 1 & $\sim60$ \\ \hline
        3 & Intel RealSense D435i Camera & 1 & $\sim346$ \\ \hline
        4 & Logitech BRIO & 1 & $\sim150$ \\ \hline
        5 & RØDE VideoMicro II & 1 & $\sim60$ \\ \hline

        6 & Protective Grille for $80x80mm$ Fan & 1 & $\sim1.30$ \\ \hline
        7 & Ethernet Cable Slim $1m$ & 1 & $\sim1$  \\ \hline
        8 & USB Type-C Cable 5 Gbit/s $1m$ & 1 & $\sim5$ \\ \hline
        9 & USB Type-C Cable 5 Gbit/s $0.25m$ & 1 & $\sim5$ \\ \hline

        10 & 90° L/R Adapter USB 3.1 C/C & 1 & $\sim3$ \\ \hline
        11 & External USB sound card & 1 & $\sim15$ \\ \hline
        12 & 3.5 mm TRS to TRS cable $1m$ & 1 & $\sim2$ \\ \hline
    
        13 & USB Male to Male Y-Splitter & 1 & $\sim5.5$ \\ \hline
        14 & DC/DC-Converter R-78B15-2.0 & 1 & $\sim12$ \\ \hline

        15 & Filament (PLA) & $\sim1kg$ & $\sim26$ \\ \hline
        16 & Filament (PCTG) & $\sim0.8kg$ & $\sim30$ \\ \hline
        17 & 3D Printing Thread Inserts & 22 & $\sim3$ \\ \hline
        18 & Hex screw M3, $8mm$ & 14 & $\sim1.4$ \\ \hline
        19 & Hex screw M3, $10mm$ & 8 & $\sim0.8$ \\ \hline
        20 & Hex screw M3, $20mm$ & 2 & $\sim0.2$ \\ \hline
        21 & Hex screw M5, $16mm$ & 1 & $\sim0.2$ \\ \hline
        22 & Hex screw M6, $12mm$ & 1 & $\sim0.2$ \\ \hline
        23 & Hex Nut M5 & 1 & $\sim0.2$ \\
        \midrule
        \multicolumn{3}{l|}{\textbf{Total}} & $\sim$\totalcost \\ \bottomrule
    \end{tabular}
\end{table}

\subsection{Mechanical and Hardware Validation}
\begin{figure}[htbp]
    \centering
    \includegraphics[width=0.3\linewidth,trim=0 1050 0 0,clip]{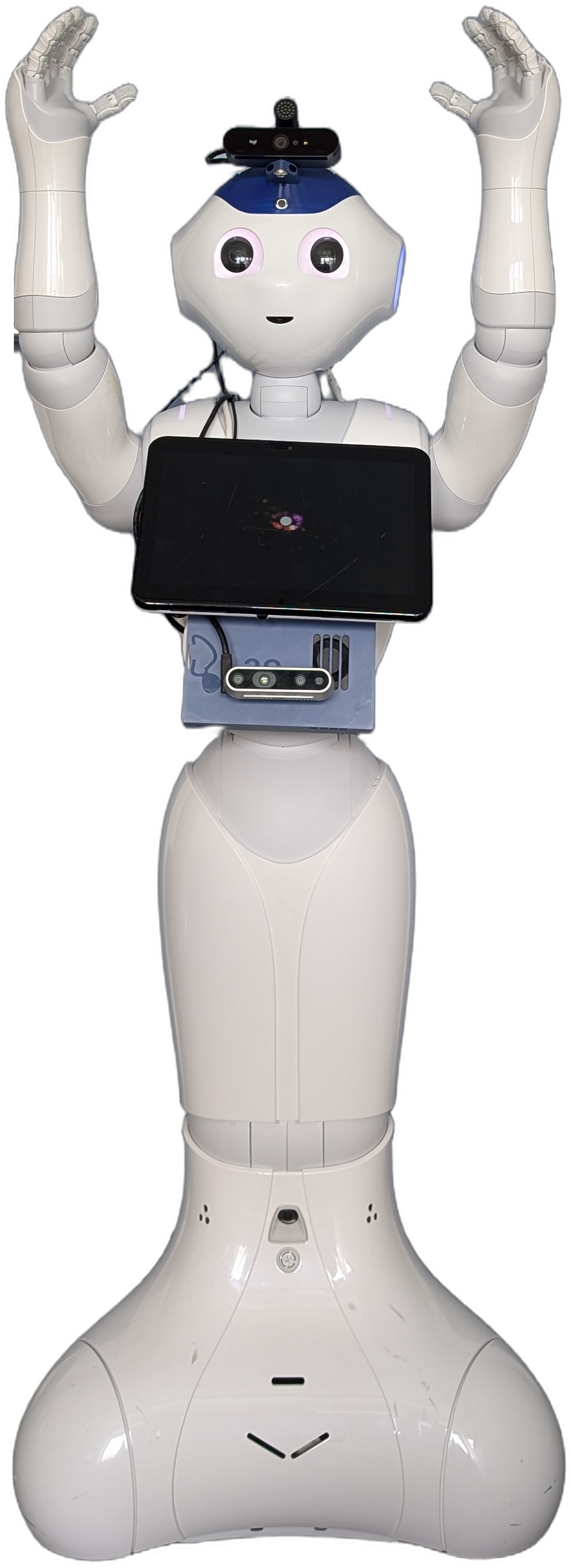}
    \includegraphics[width=0.3\linewidth,trim=0 970 0 0,clip]{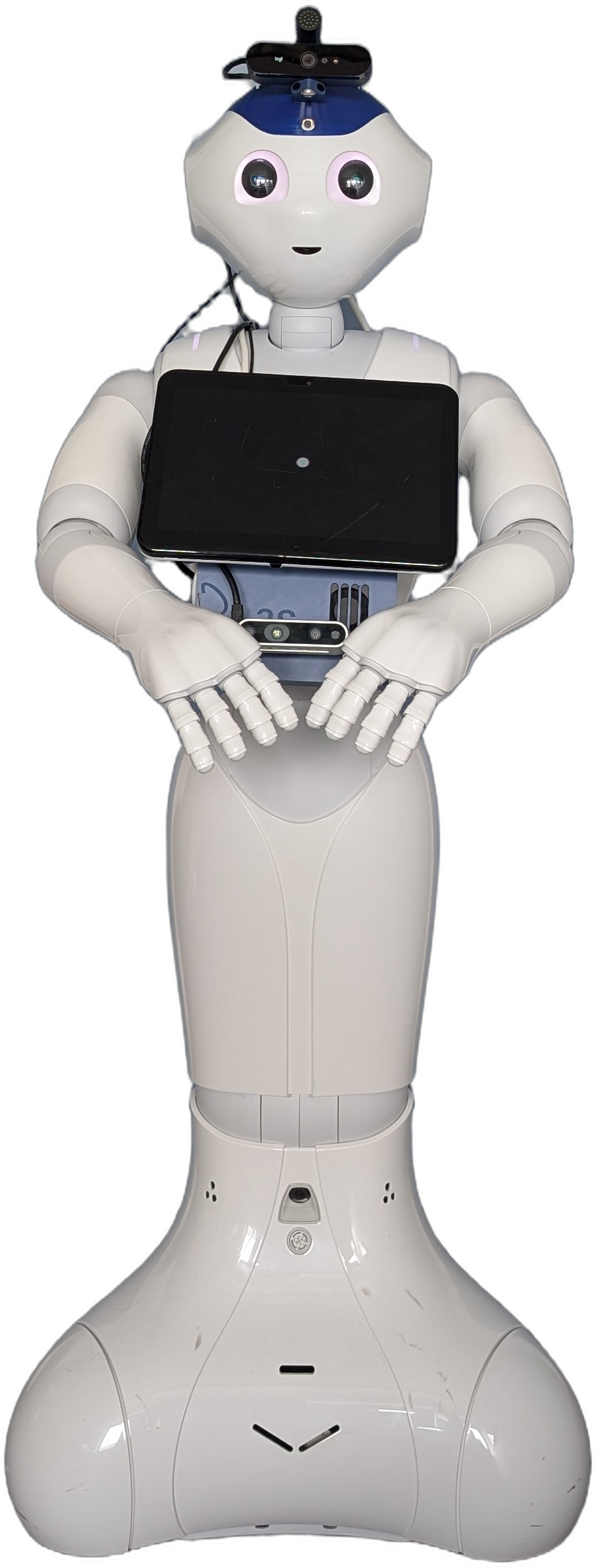}
    \includegraphics[width=0.3\linewidth,trim=0 920 0 0,clip]{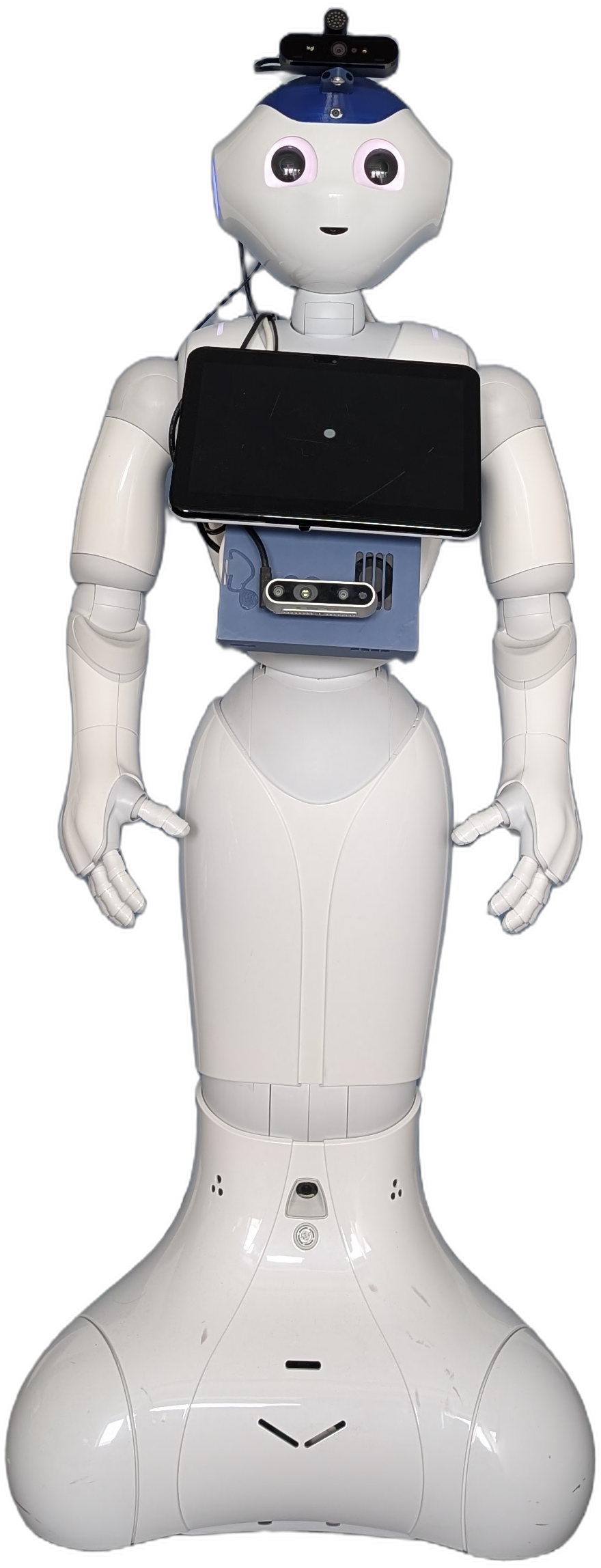}

    \includegraphics[width=0.265\linewidth,trim=0 1100 0 0,clip]{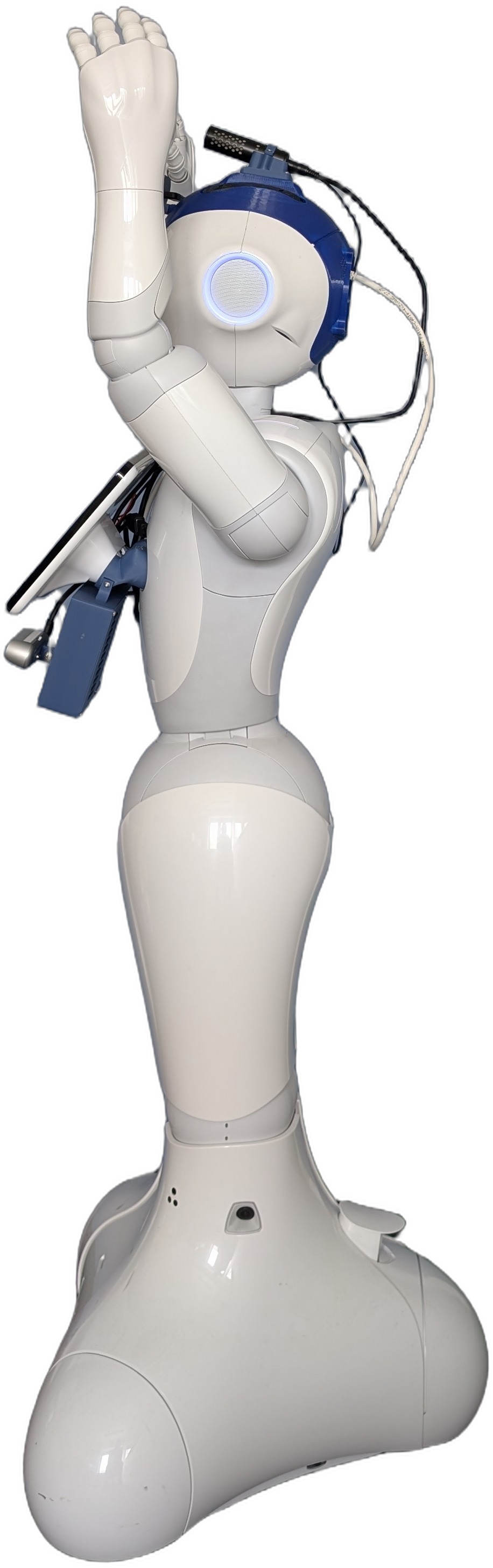}
    \includegraphics[width=0.35\linewidth,trim=0 920 0 0,clip]{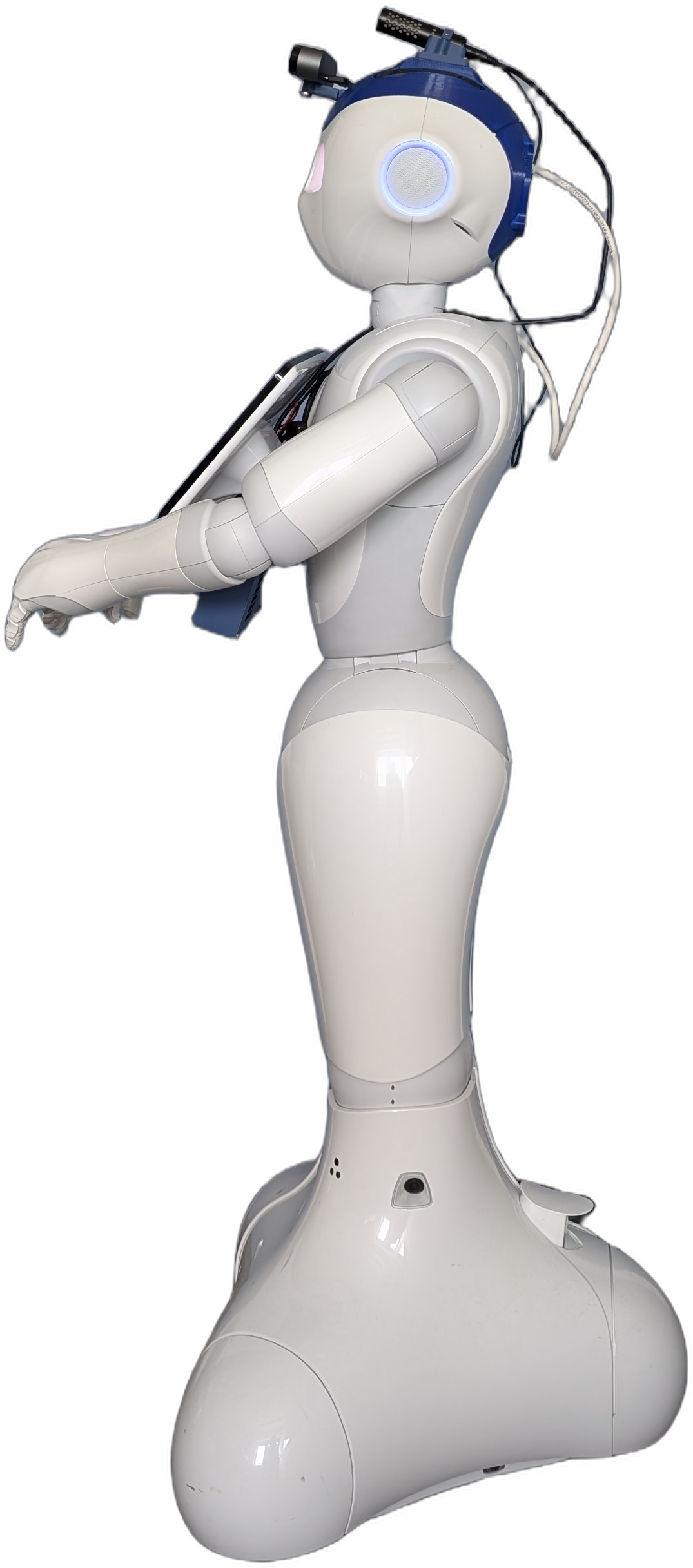}
    \includegraphics[width=0.265\linewidth,trim=0 920 0 0,clip]{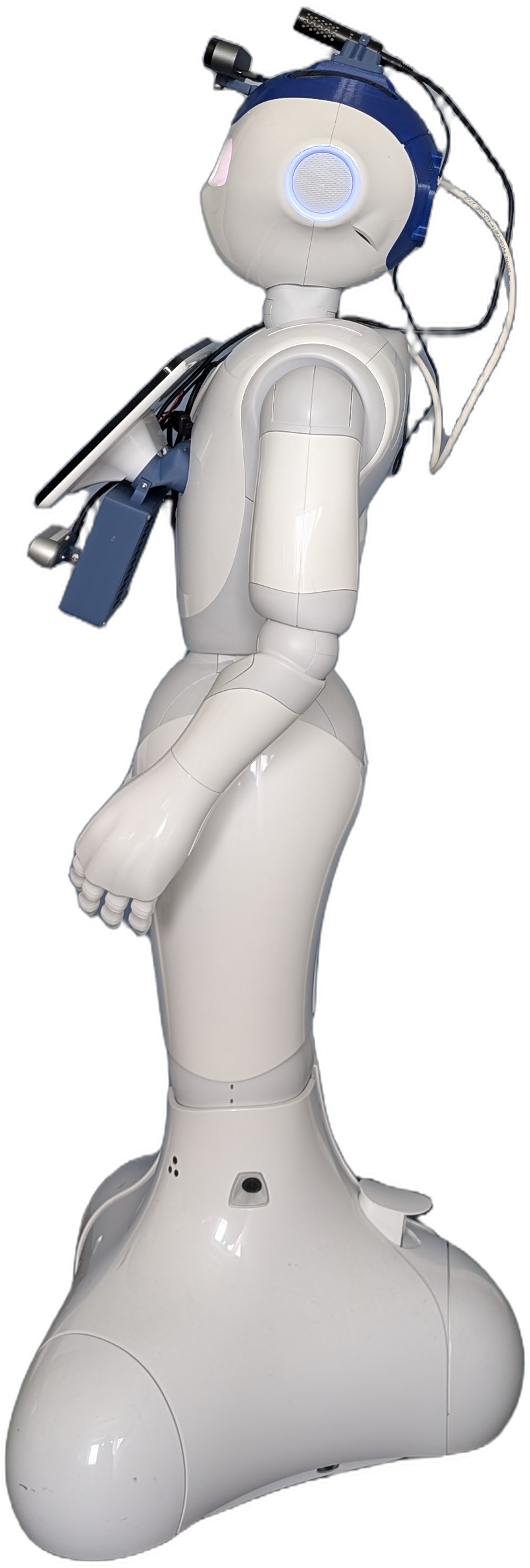}
    \caption{Assembled version of {\addonname}. Demonstration of arm mobility.}
    \label{fig:motion_clearance}
\end{figure}

Mechanical verification confirmed that the 3D-printed covers, holders, and the Jetson enclosure fit securely on Pepper and do not interfere with the motion of the head, arms, or tablet.  
Fig.~\ref{fig:motion_clearance} shows the assembled version of the {\addonname} and a sequence of arm movements demonstrating the range of motion which is slightly limited below the Tablet. However, the startup sequence of Pepper is not obstructed. 

\subsection{Thermal Validation}
\begin{figure}[htbp]
    \centering
    \includegraphics[width=\linewidth]{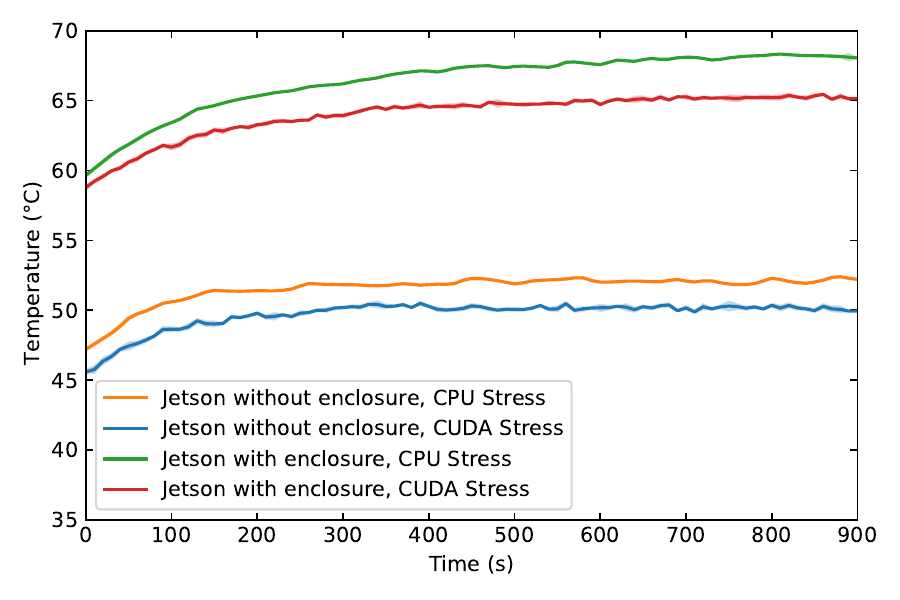}
    \includegraphics[width=\linewidth]{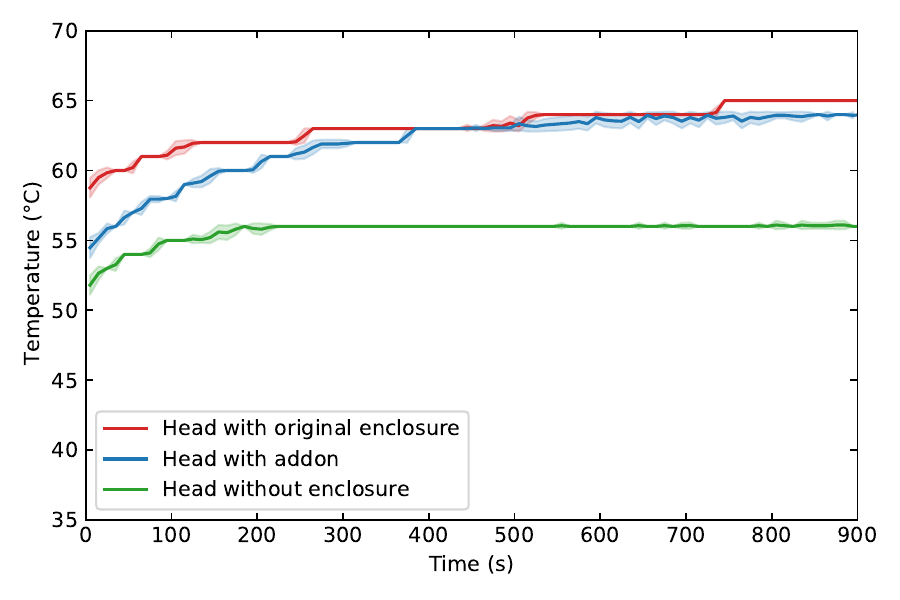}
    \caption{Thermal performance under high-load conditions. 
    Top: temperature evolution of the Jetson Orin Nano during CPU and CUDA-bound inference workloads in open-air and enclosed add-on configurations. 
    Bottom: temperature evolution of Pepper’s head CPU during a 100\% utilization test comparing the head without enclosure, with the original enclosure, and with the add-on enclosure.}
    \label{fig:stress_test}
\end{figure}

Thermal validation results are shown in Fig.~\ref{fig:stress_test} for both the Jetson~Orin~Nano and Pepper’s internal processor.  
For each platform, temperature was recorded continuously during high-load operation across the tested configurations described in Section~\ref{sec:methodology}.  

The figure presents the recorded temperature profiles for both systems: the upper plot shows the Jetson~Orin~Nano under CPU and CUDA-bound inference workloads in the open-air and enclosed configurations. In contrast, the lower plot shows Pepper’s head CPU during full-load operation in the three enclosure configurations.

It is important to note the manufacturer-rated thermal limits for each processing platform. 
The Jetson~Orin~Nano module specifies an operational junction temperature limit of up to 90\,°C \cite{nvidia_orin_nano_ds}, 
whereas Pepper's onboard processor (Intel Atom~E3845) supports junction temperatures up to approximately 110\,°C \cite{pepper_specs,intel_e3845_thermal}.

\subsection{Wireless Interfaces}
Table~\ref{tab:wireless_perf} reports the measured wireless performance of the Pepper robot and the Jetson Orin Nano platform at distances from 1\,m to 32\,m under line-of-sight (LoS) and obstructed (Obs) conditions. 

For each distance and condition, the received signal strength indicator was recorded at the platform when available, and the data-transfer time was measured for sending dummy files of different sizes (\texttt{.txt} 0.01 MB, \texttt{.wav} 2 MB, \texttt{.png} 5 MB, and \texttt{.bin} 50 MB) over Wi-Fi.

The table lists mean $\pm$ standard-deviation values for all tested combinations.  
For the Jetson, the RSSI decreases from \(-33\,\mathrm{dBm}\) at 1\,m under line-of-sight to \(-84\,\mathrm{dBm}\) at 32\,m under obstructed conditions.  
Pepper’s internal Wi-Fi does not report RSSI values.  

Data-transfer times for all file types generally increase with distance and further rise under obstructed conditions.  
At 1–16\,m, both platforms complete transfers for all file types, although times are consistently higher under obstruction.  
At 32\,m with obstruction, Pepper could not transfer any file because the robot was unable to establish a wireless connection.

\begin{table*}[htbp]
    \centering
    \small
    \renewcommand{\arraystretch}{1.15}
    \setlength{\tabcolsep}{4pt}
    \caption{Wireless network–interface performance of the Pepper robot and Jetson Orin Nano platform at different distances under line-of-sight (LoS) and obstructed (Obs) conditions. 
    Values are reported as mean$\pm$std. RSSI is the received signal strength at the platform  (not available for Pepper’s built-in Wi-Fi). Data-transfer time is measured for sending dummy files (\texttt{.txt} 0.01MB, \texttt{.wav} 2MB, \texttt{.png} 5MB, \texttt{.bin} 50MB) from each platform over Wi-Fi.}
    \label{tab:wireless_perf}

    \begin{tabular}{@{}clc cc cc@{}}
    \toprule
     &  &  & \multicolumn{2}{c}{\textbf{Jetson}} & \multicolumn{2}{c}{\textbf{Pepper}} \\
    \cmidrule(lr){4-5}\cmidrule(lr){6-7}
    \textbf{Dist.\ (m)} & \multicolumn{2}{c}{\textbf{Metric}} & \textbf{LoS} & \textbf{Obs} & \textbf{LoS} & \textbf{Obs} \\
    \midrule
    
    \multirow{5}{*}{1} & \multicolumn{2}{l}{RSSI (dBm)} & $-33 \pm 0$ & $-54 \pm 0$ & N/A & N/A \\
     & \multirow{4}{*}{Data transfer time (s)} & Text  & $0.41 \pm 0.16$ & $0.52 \pm 0.06$ & $0.38 \pm 0.06$ & $0.57 \pm 0.05$ \\
     & & Audio & $0.77 \pm 1.03$ & $0.98 \pm 0.49$ & $1.51 \pm 0.13$ & $0.83 \pm 1.65$ \\
     & & Image & $1.29 \pm 0.05$ & $0.65 \pm 0.64$ & $3.44 \pm 0.50$ & $4.29 \pm 2.00$ \\
     & & LIDAR & $9.08 \pm 0.39$ & $17.06 \pm 0.74$ & $17.29 \pm 1.14$ & $23.36 \pm 1.38$ \\
    \midrule
    
    \multirow{5}{*}{2} & \multicolumn{2}{l}{RSSI (dBm)} & $-35 \pm 0$ & $-55 \pm 0$ & N/A & N/A \\
     & \multirow{4}{*}{Data transfer time (s)} & Text  & $0.41 \pm 0.11$ & $0.60 \pm 0.10$ & $0.39 \pm 0.04$ & $0.56 \pm 0.05$ \\
     & & Audio & $0.76 \pm 0.44$ & $1.23 \pm 0.05$ & $1.44 \pm 0.10$ & $0.99 \pm 0.43$ \\
     & & Image & $1.25 \pm 0.03$ & $0.89 \pm 0.74$ & $3.22 \pm 0.49$ & $4.68 \pm 1.97$ \\
     & & LIDAR & $8.77 \pm 0.33$ & $17.36 \pm 0.85$ & $17.04 \pm 1.11$ & $23.97 \pm 1.36$ \\
    \midrule
    
    \multirow{5}{*}{4} & \multicolumn{2}{l}{RSSI (dBm)} & $-40 \pm 0$ & $-57 \pm 0$ & N/A & N/A \\
     & \multirow{4}{*}{Data transfer time (s)} & Text  & $0.40 \pm 0.09$ & $0.65 \pm 0.13$ & $0.41 \pm 0.44$ & $0.56 \pm 0.05$ \\
     & & Audio & $0.73 \pm 0.35$ & $1.36 \pm 0.31$ & $1.32 \pm 0.04$ & $1.32 \pm 1.48$ \\
     & & Image & $1.22 \pm 2.62$ & $1.44 \pm 0.94$ & $2.79 \pm 0.47$ & $5.44 \pm 1.91$ \\
     & & LIDAR & $8.33 \pm 0.22$ & $18.11 \pm 1.07$ & $16.35 \pm 1.05$ & $26.37 \pm 1.31$ \\
    \midrule
    
    \multirow{5}{*}{8} & \multicolumn{2}{l}{RSSI (dBm)} & $-45 \pm 0$ & $-61 \pm 0$ & N/A & N/A \\
     & \multirow{4}{*}{Data transfer time (s)} & Text  & $0.42 \pm 0.10$ & $0.67 \pm 0.14$ & $0.41 \pm 0.03$ & $0.58 \pm 0.06$ \\
     & & Audio & $0.78 \pm 1.85$ & $1.44 \pm 0.44$ & $1.32 \pm 0.05$ & $1.98 \pm 0.13$ \\
     & & Image & $1.33 \pm 0.03$ & $2.44 \pm 1.32$ & $2.60 \pm 0.10$ & $7.03 \pm 1.77$ \\
     & & LIDAR & $9.23 \pm 0.18$ & $19.96 \pm 1.56$ & $15.49 \pm 1.44$ & $30.82 \pm 1.24$ \\
    \midrule
    
    \multirow{5}{*}{16} & \multicolumn{2}{l}{RSSI (dBm)} & $-56 \pm 0$ & $-69 \pm 0$ & N/A & N/A \\
     & \multirow{4}{*}{Data transfer time (s)} & Text  & $0.39 \pm 0.34$ & $0.68 \pm 0.14$ & $0.40 \pm 0.14$ & $0.59 \pm 0.08$ \\
     & & Audio & $0.73 \pm 0.11$ & $1.47 \pm 0.50$ & $1.58 \pm 0.31$ & $3.29 \pm 2.05$ \\
     & & Image & $1.18 \pm 0.07$ & $4.49 \pm 2.10$ & $2.64 \pm 0.11$ & $10.18 \pm 1.49$ \\
     & & LIDAR & $9.01 \pm 2.20$ & $23.52 \pm 2.52$ & $15.68 \pm 1.62$ & $39.79 \pm 1.07$ \\
    \midrule
    
    \multirow{5}{*}{32} & \multicolumn{2}{l}{RSSI (dBm)} & $-49 \pm 0$ & $-84 \pm 0$ & N/A & N/A \\
     & \multirow{4}{*}{Data transfer time (s)} & Text  & $0.88 \pm 2.15$ & $0.70 \pm 0.15$ & $0.41 \pm 0.27$ & No signal \\
     & & Audio & $0.75 \pm 0.04$ & $1.47 \pm 0.53$ & $1.58 \pm 0.14$ & No signal \\
     & & Image & $1.27 \pm 0.09$ & $8.64 \pm 3.65$ & $3.02 \pm 0.19$ & No signal \\
     & & LIDAR & $8.78 \pm 0.24$ & $31.02 \pm 4.33$ & $17.80 \pm 1.67$ & No signal \\
    \bottomrule
    \end{tabular}
\end{table*}

\subsection{Microphones}
The acoustic evaluation compares Pepper's front-facing built-in microphone with two external devices: the Samson UB1 (omnidirectional) and the RØDE VideoMicro II (directional).  
Measurements were taken at a distance of 1.0 m from a broadband sound source, under three angular orientations: 0°, 45°, and 90° relative to the source.

\begin{figure*}[htbp]
    \centering
        \includegraphics[width=0.3275\textwidth]{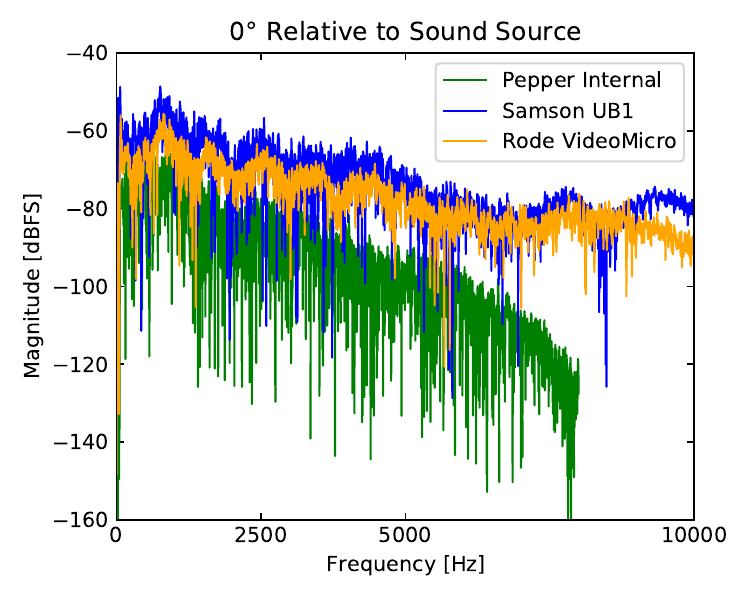}
        \includegraphics[width=0.3275\textwidth]{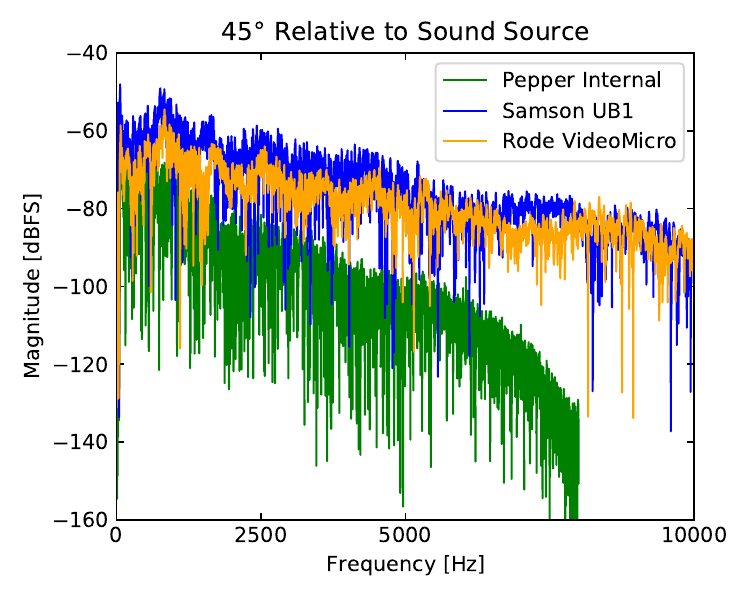}
        \includegraphics[width=0.3275\textwidth]{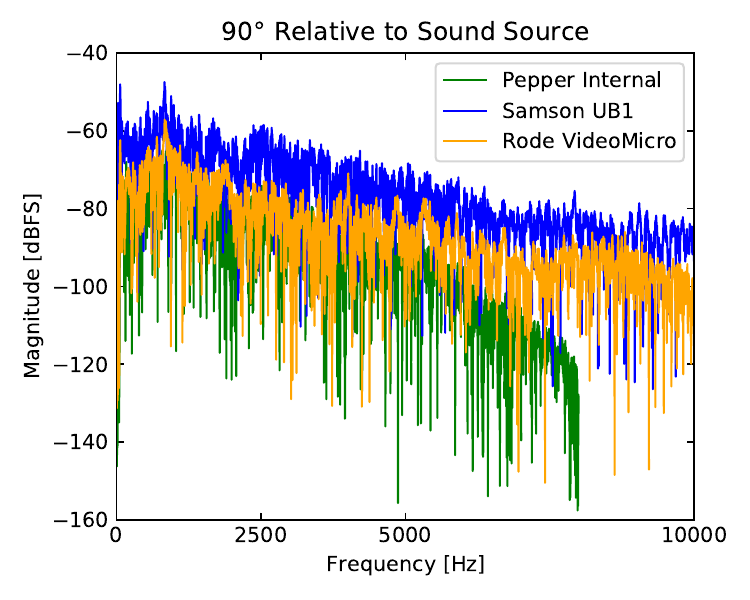}
    
    \caption{Frequency response of the tested microphones at 1.0\,m from the sound source, measured at orientations of 0\textdegree, 45\textdegree, and 90\textdegree. The curves correspond to the Samson UB1 (blue), the RØDE VideoMicro II (orange), and the Pepper internal microphone (green).}
    \label{fig:mic_response}
\end{figure*}

Fig.~\ref{fig:mic_response} shows the corresponding frequency responses of the three tested microphones, reporting the magnitude in dB (SPL) as a function of frequency.  
The Pepper microphone exhibits lower noise floors and reduced sensitivity at mid-to-high frequencies, particularly above 2 kHz.  
The Samson UB1 provides the most stable response across angles, whereas the RØDE VideoMicro II shows a slightly stronger on-axis response but reduced gain at 90°, consistent with its directional design.

Table~\ref{tab:mic_snr} reports the signal-to-noise ratio within the 300–3400 Hz band.  
Although the measurements covered the full 0–10 kHz range, this band is highlighted because it is the most relevant for human speech perception.
The Pepper microphone exhibits the lowest SNR across all orientations (ca.\ 5 dB), while both external microphones display higher values (ca.\ 14–16 dB).

\begin{table}[htbp]
    \centering
    \renewcommand{\arraystretch}{1.2}
    \setlength{\tabcolsep}{3pt}
    \caption{Signal-to-noise ratio (SNR) of tested microphones at 1.0 m (Band 300–3400 Hz).}
    \label{tab:mic_snr}
    \begin{tabular}{@{}l|ccc@{}}
        \toprule
        \textbf{Microphone} & \textbf{0°} & \textbf{45°} & \textbf{90°} \\
        \midrule
        Pepper Internal & 4.7 $\pm$ 8.5 & 4.8 $\pm$ 8.9 & 5.0 $\pm$ 9.2 \\
        Samson UB1 & 14.6 $\pm$ 23.0 & 15.1 $\pm$ 22.9 & 15.3 $\pm$ 22.2 \\
        RØDE VideoMicro & 15.4 $\pm$ 23.7 & 15.8 $\pm$ 23.5 & 14.5 $\pm$ 21.5 \\
        \bottomrule
    \end{tabular}
\end{table}

\subsection{RGB Cameras}
Table \ref{tab:rgb_metrics} reports the aggregated metrics for the RGB images acquired with Pepper’s internal color camera and the Intel RealSense D435i color frames.
The table includes the mean and standard deviation of luminance over the 8-bit range, the luminance entropy in bits, the fractions of under-exposed and over-exposed pixels, the Laplacian-based sharpness, the Hasler–Süsstrunk colorfulness index, and the total number of frames analyzed.

\begin{table*}
  \centering
  \caption{RGB stream quality metrics.}
  \label{tab:rgb_metrics}
  \begin{tabular}{@{}lrrr@{}}
    \toprule
    \textbf{Metric} & \textbf{Pepper RGB} & \textbf{RealSense RGB} & \textbf{BRIO RGB} \\
    \midrule
    Resolution (px) & 1280$\times$960 & 1920$\times$1080 & 3840$\times$2160 \\
    FPS & 3.86 & 15.96 & 27.07 \\
    Mean luminance (0--255) & 141.51 & 114.04 & 123.85 \\
    Luminance std.\ dev.\ (0--255) & 44.69 & 38.05 & 37.77 \\
    Luminance entropy (bits) & 7.37 & 6.88 & 6.90 \\
    Under-exposed pixels (\%) & 0.17 & 0.00 & 0.75 \\
    Over-exposed pixels (\%) & 0.71 & 0.04 & 0.05 \\
    Sharpness (Laplacian variance) & 176.47 & 57.90 & 89.22 \\
    Colorfulness (Hasler--S\"usstrunk) & 24.91 & 13.10 & 18.05 \\
    Number of frames & 117 & 482 & 819 \\
    \bottomrule
  \end{tabular}
\end{table*}

Additionally, we report the accuracy during face recognition based on the quality of the images. The results can be found in Table~\ref{tab:face_recognition}. Overall, when utilizing Max Resolution, the BRIO camera achieves $64.4\%$ accuracy for face detection and $60\%$ for face recognition, while the Pepper internal camera achieves $31.1\%$ and $28.3\%$ respectively. 
For Max FPS, the BRIO camera achieves $60.6\%$ for face detection and $48.3\%$ for face recognition. Pepper achieves $0.0\%$ accuracy for both metrics.

The BRIO camera exceeds the Pepper internal camera in all categories, except \textbf{Overexposure}, which resulted in a $0.0\%$ detection and recognition rate in both cases. Pepper cannot perform at the max FPS settings. The Max Resolution setting for the BRIO camera led to higher scores than under the Max FPS setting, except in the \textbf{Human head movement} and \textbf{Pepper head movement}.

\begin{table*}[htbp]
    \centering
    \caption{Table reports the accuracy rates during face recognition between Peppers internal camera and the BRIO camera.}
    \label{tab:face_recognition}
    \small
    \setlength{\tabcolsep}{4pt} 
    \begin{tabular}{@{}c|cc|cc|cc|cc@{}}
          \toprule
             & \multicolumn{4}{c|}{Max Resolution} & \multicolumn{4}{c}{Max FPS} \\ \midrule
            \textbf{Condition} & \multicolumn{2}{c|}{\textbf{Pepper}} & \multicolumn{2}{c|}{\textbf{BRIO}} & \multicolumn{2}{c|}{\textbf{Pepper}} & \multicolumn{2}{c}{\textbf{BRIO}} \\ \midrule
             & Det.\ & Rec.\ & Det.\ & Rec.\ & Det.\ & Rec.\ & Det.\ & Rec.\ \\ \midrule
            \textbf{Frontal} & $96.7\%$ & $93.3\%$ & $100.0\%$ & $96.7\%$ & $0.0\%$ & $0.0\%$ & $100.0\%$ & $86.7\%$\\
            \textbf{Human head movement} & $73.3\%$ & $60.0\%$ & $73.3\%$ & $70.0\%$ & $0.0\%$ & $0.0\%$ & $80.0\%$ & $76.7\%$\\
            \textbf{Pepper head movement} & $0.0\%$ & $0.0\%$ & $30.0\%$ & $26.7\%$ & $0.0\%$ & $0.0\%$ & $63.3\%$ & $53.3\%$\\
            \textbf{Distance} & $16.7\%$ & $16.7\%$ & $100.0\%$ & $90.0\%$ & $0.0\%$ & $0.0\%$ & $36.7\%$ & $33.3\%$\\
            \textbf{Overexposure} & $0.0\%$ & $0.0\%$ & $0.0\%$ & $0.0\%$ & $0.0\%$ & $0.0\%$ & $0.0\%$ & $0.0\%$\\
            \textbf{Low lighting} & $0.0\%$ & $0.0\%$ & $83.3\%$ & $80.0\%$ & $0.0\%$ & $0.0\%$ & $83.3\%$ & $40.0\%$\\
            \bottomrule
            \textbf{Overall} & $31.1\%$ & $28.3\%$ & $64.4\%$ & $60.6\%$ & $0.0\%$ & $0.0\%$ & $60.6\%$ & $48.3\%$\\
          \bottomrule
    \end{tabular}
\end{table*}

\subsection{Depth Cameras}
Table \ref{tab:depth_metrics} summarizes the corresponding metrics for the depth data acquired with Pepper’s internal depth frames and the Intel RealSense D435i depth frames.
It lists the mean and standard deviation of raw depth values, the entropy of the depth histogram, the proportions of valid and invalid pixels, the low- and high-saturation fractions, the mean normalized gradient magnitude, the bit depth of the stored frames, and the total number of frames processed.

\begin{table*}
  \centering
  \caption{Depth stream quality metrics.}
  \label{tab:depth_metrics}
  \begin{tabular}{lrr}
        \toprule
        \textbf{Metric} & \textbf{Pepper Depth} & \textbf{RealSense Depth} \\
        \midrule
        Resolution (px) & 1280$\times$720 & 1280$\times$720 \\
        FPS & 0.81 & 21.61 \\
        Mean depth value (mm) & 915.66 & 3076 \\
        Depth standard deviation (mm) & 1257 & 5786 \\
        Depth entropy (bits) & 2.45 & 4.27 \\
        Valid pixel ratio (\%) & 40.78 & 90.50 \\
        Invalid (zero) pixel ratio (\%) & 59.22 & 9.50 \\
        Low-saturation fraction (\%) & 59.22 & 9.50 \\
        High-saturation fraction (\%) & 0.00 & 0.34 \\
        Bit depth (bits) & 16 & 16 \\
        Number of frames & 25 & 652 \\
        \bottomrule
  \end{tabular}
\end{table*}

\begin{figure}[ht]
    \centering
    \includegraphics[width=0.85\linewidth]{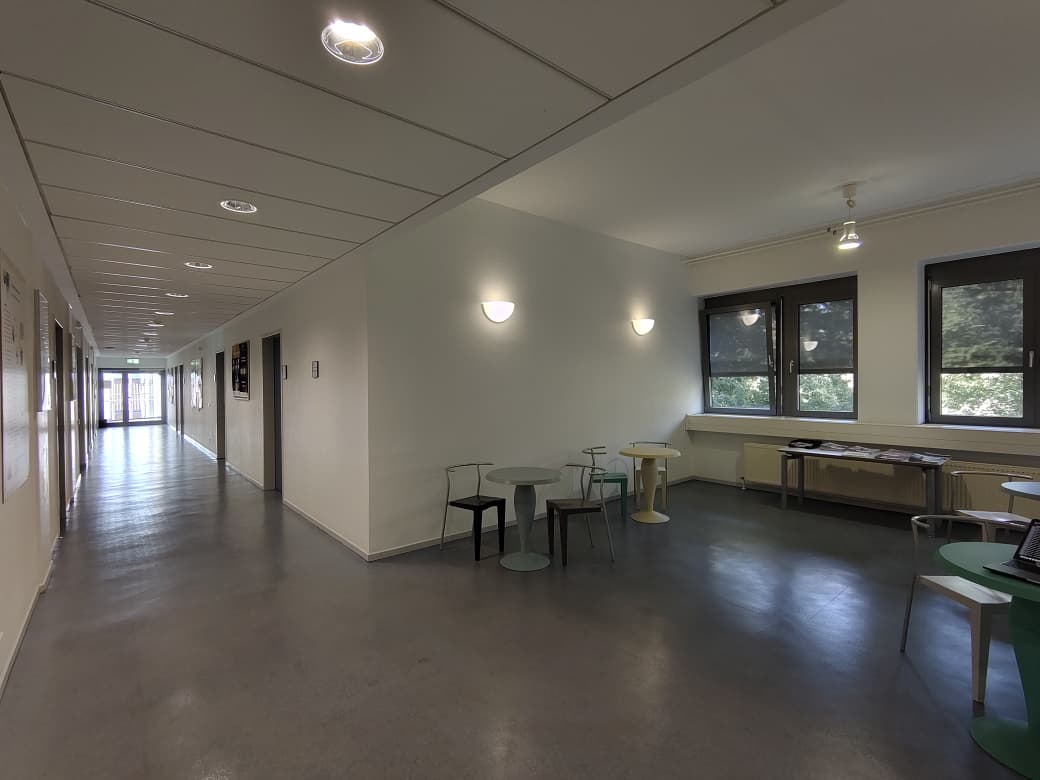}
    \caption{Photograph of the indoor mapping site: a straight corridor with an adjacent lounge area furnished with tables and chairs, illuminated by ceiling lights and windows.}
    \label{fig:scene_mapping}
\end{figure}

To compare the perception capabilities of the baseline Pepper configuration and {\addonname} equipped with Jetson and D435i, both were evaluated on a representative indoor mapping task.  
Pepper was teleoperated along a predefined trajectory that covered the full length of the corridor and entered the lounge. 
The head pose was kept fixed during acquisition to avoid additional extrinsic variation. 
Fig.~\ref{fig:scene_mapping} shows a picture of the test area, illustrating the straight corridor, the adjacent lounge with tables and chairs, artificial lights, and natural illumination from windows.  

\begin{figure*}
    \centering
    \includegraphics[width=1\linewidth]{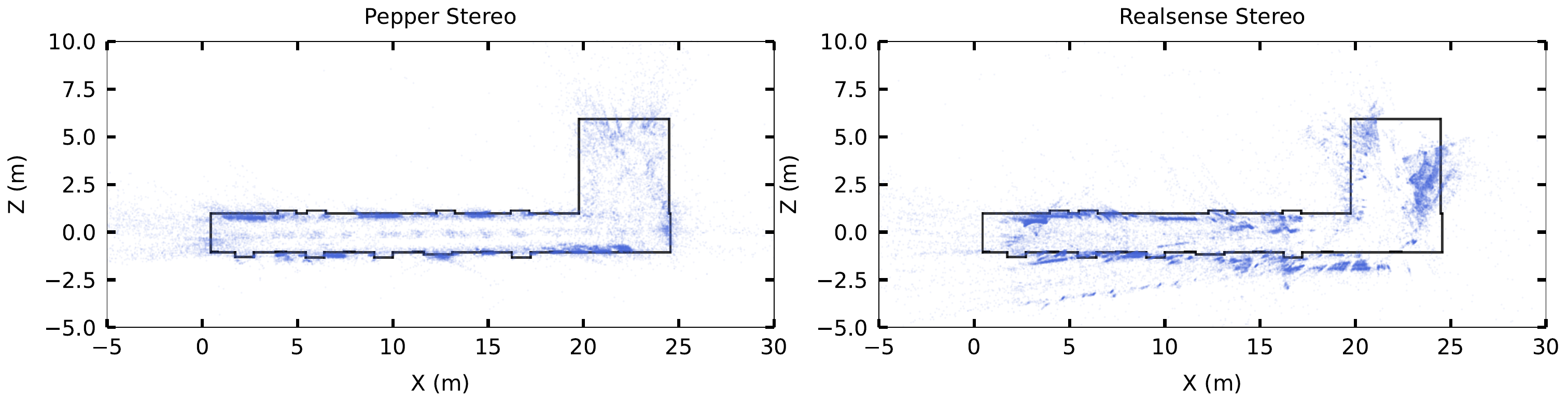}
    \caption{Top-Down Indoor mapping results overlapped with ground truth schematics for Stereo mapping with ORB-SLAM3 (Atlas Maps).}
    \label{fig:stereo}
\end{figure*}

\begin{figure*}
    \centering
    \includegraphics[width=1\linewidth]{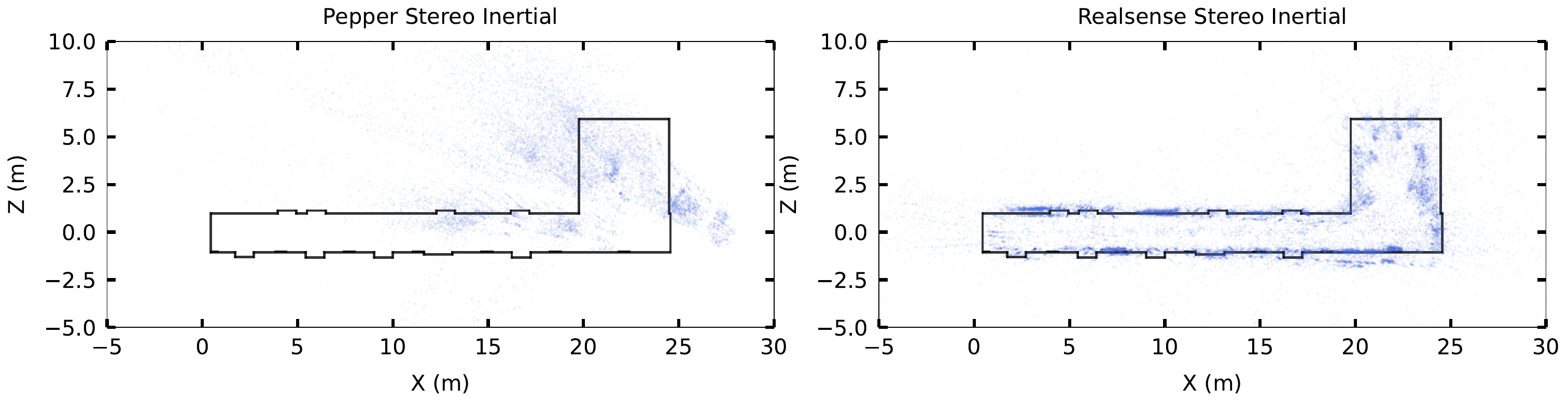}
    \caption{Top-Down Indoor mapping results overlapped with ground truth schematics for Stereo+IMU mapping with ORB-SLAM3 (Atlas Maps).}
    \label{fig:stereo-inertial}
\end{figure*}

Fig.\ \ref{fig:stereo} and Fig.\ \ref{fig:stereo-inertial} show the atlas maps generated by ORB-SLAM3 for the same dataset, where Fig.\ \ref{fig:stereo} uses stereo data alone and Fig.\ \ref{fig:stereo-inertial} incorporates both stereo and IMU data.

Qualitatively, the Pepper internal camera produced better maps in stereo-only mode, while the Intel RealSense D435i achieved the best results when stereo was combined with IMU data. The Stereo+IMU setup was tested, and it is recommended as it provides robustness by helping  improve motion estimation during fast motions, visual dropouts, and rotations in low-texture scenes.

\section{Analysis and Discussion}
\label{sec:discussion}

\subsection{Mechanical and thermal behavior}
The add-on mounts cleanly and preserves Pepper’s kinematic envelope (Fig.~\ref{fig:motion_clearance}).
Under sustained inference, the Jetson Orin Nano exhibits the expected temperature increase when enclosed, yet the rise remains modest and workload-dependent (CUDA $>$ CPU), indicating that enclosure airflow is adequate (Fig.~\ref{fig:stress_test}, top).
Pepper’s head CPU shows similar or slightly \emph{lower} steady-state temperatures with the add-on enclosure compared to the stock shell (Fig.~\ref{fig:stress_test}, bottom), suggesting that the modified head covers do not impede cooling.
These results support continuous on-board inference on the Jetson without compromising the robot’s native thermal budget.

\subsection{Wireless connectivity}
Across distances and through obstruction, the Jetson’s Wi-Fi interface sustains stronger RSSI and faster, more stable transfers than Pepper’s built-in adapter (Table~\ref{tab:wireless_perf}).
Through-wall conditions magnify this gap: file-transfer times increase for both platforms, but the Jetson remains operational at 32~m whereas Pepper fails to associate.
For teleoperation, log offloading, and remote model deployment, the add-on therefore increases the practical operating envelope and reduces transfer-time variability indoors.

\subsection{Audio perception}
External microphones materially improve far-field capture relative to Pepper’s internal microphone frequency responses at 0°, 45°, and 90° show lower noise floors and better mid–high sensitivity for both Samson~UB1 and RØDE~VideoMicro~II (Fig.~\ref{fig:mic_response}).
In the speech band (300–3400~Hz), SNR increases from $\approx$5~dB for the internal microphone and to $\approx$15~dB for the external ones (Table~\ref{tab:mic_snr}).
The Samson~UB1’s angle-invariant response favors unconstrained interaction, while the RØDE~VideoMicro~II’s on-axis gain and off-axis attenuation are advantageous for selective capture and ambient-noise rejection.
Practically, these characteristics should reduce false wakeups and improve ASR robustness in open spaces.

\subsection{RGB Cameras}
The consistent superiority of the BRIO camera is attributable to its higher resolution, which preserves finer facial detail and thereby improves both detection and recognition reliability. The advantage of the BRIO's Max~FPS mode over Max~Resolution mode in the \textbf{Human head movement} and \textbf{Pepper head movement} scenarios is similarly explained by the effectively shorter exposure time at higher frame rates, which reduces motion blur on moving subjects.

The \textbf{Overexposure} scenario represents a consistent failure case across both cameras, yielding $0.0\%$ accuracy in all settings. Manual inspection confirmed that the strong backlighting from a window fully saturated the sensor, rendering faces undetectable. This outcome reflects a fundamental limitation of passive RGB sensing under adverse illumination rather than a camera-specific deficiency, and motivates the inclusion of active depth sensing for robustness in unconstrained environments.

\subsection{Stereo Cameras and Mapping}
Qualitatively, the Pepper internal camera produced superior maps when using stereo data alone, whereas the Intel RealSense D435i camera yielded better results when fusing stereo with IMU data. We attribute the degraded stereo-only performance of the Intel RealSense D435i to its mounting position, which places it closer to nearby objects such as chairs and tables than the Pepper internal camera. This reduced working distance appears to adversely affect feature tracking and depth estimation in ORB-SLAM3.

In the Stereo+IMU configuration, the Pepper camera exhibited difficulties in effectively incorporating IMU measurements into the mapping process. We attribute this to a substantial temporal offset between the IMU data and stereo frames delivered by the Pepper platform, which likely disrupts the tightly-coupled visual-inertial estimator and degrades trajectory consistency.

Overall, the Intel RealSense D435i camera operating in Stereo+IMU mode achieved the highest qualitative map quality, successfully reconstructing the full layout of the waiting area including furniture such as tables and chairs, which are visible as the point concentrations inside the waiting area.

\subsection{Comparative Evaluation}
During the comparative evaluation, none of the cameras achieved their theoretical maximum frame rate. For the Logitech BRIO and Intel RealSense D435i, we attribute this shortfall to the reduced power mode used in this implementation; however, higher power draws can be achieved using a different DC/DC converter while still remaining under the suggested limit~\cite{Caniot2020Adapted}. For Pepper's built-in cameras, the lower frame rate is attributable to remote frame streaming, which is documented to reduce the maximum achievable FPS.

\subsubsection{RGB Quality Metrics}
All three RGB sensors produce well-exposed images under the tested indoor conditions, with mean luminance values ranging from 114 to 142 out of 255. The Intel RealSense D435i achieves near-perfect exposure control, with zero under-exposed pixels across all frames, while the BRIO exhibits mild auto-exposure fluctuation ($0.75\%$ under-exposed pixels; standard deviation of $0.0023$ across frames), suggesting that exposure had not fully settled during the capture session. Pepper shows slight highlight clipping at $0.71\%$ over-exposed pixels, consistent with its higher mean luminance.

To eliminate resolution bias in the Laplacian variance metric, we normalised all images to a common $640\times480$ reference resolution before computing sharpness. Under this normalisation, Pepper scores highest (176.5), followed by BRIO (89.2) and Intel RealSense D435i (57.9). Notably, BRIO's sharpness standard deviation (21.1) is disproportionately large relative to its mean, yielding a coefficient of variation of $24\%$, compared with under $4\%$ for the other two cameras, indicating intermittent blur likely caused by autofocus hunting or motion artefacts during the 4K capture. The lower sharpness score of the Intel RealSense D435i reflects its design as an auxiliary sensor on a depth-centric module, where the optics are optimised for depth accuracy rather than colour image quality. Colourfulness follows the same ranking: Pepper (24.9) and BRIO (18.1) capture a more chromatically varied scene than the Intel RealSense D435i (13.1), consistent with the wider fields of view of the former two sensors.

\subsubsection{Depth Quality Metrics}
The Intel RealSense D435i depth stream substantially outperforms Pepper's across all depth quality indicators. Valid pixel coverage reaches $90.5\%$ for the Intel RealSense D435i versus only $40.8\%$ for Pepper, meaning that the majority of Pepper's depth frames contain no usable range information. The higher depth entropy of the Intel RealSense D435i ($4.27$\,bits vs.\ $2.45$\,bits) further confirms a richer scene representation, as Pepper's sparse coverage concentrates the depth histogram into fewer populated bins.

\subsection{Overall impact}
Taken together, these findings indicate that {\addonname} extends Pepper’s effective sensing and compute envelope:
\begin{inparaenum}[(i)]
    \item stable thermals under sustained inference; 
    \item substantially higher speech-band SNR with microphone choice matched to scenario requirements;
    \item qualitatively superior mapping consistent with sensor/compute advantages; and
    \item more resilient networking for data-intensive workflows.
\end{inparaenum}
Coupled with the layered software stack, these gains translate into faster prototyping on-device (Jetson-only) and seamless escalation to heavier models (laptop/cluster) without altering the mechanical integration.

\section{Conclusions and Future Work}
\label{sec:conclusions}
This project revitalizes Pepper, a widely adopted standard platform in the social robotics and HRI communities that risks obsolescence due to its limited computational power and outdated sensors. By providing an open-hardware solution, we extend Pepper's useful lifespan, enabling practitioners, developers, and researchers to continue leveraging its familiar, human-friendly design for cutting-edge applications without needing to migrate to entirely new hardware.

We approached this by designing multiple new parts for Pepper, enabling the use of better hardware. Our design complemented Pepper with a Jetson Orin Nano, a Samson UB1, a RØDE VideoMicro II, a Logitech BRIO, and an Intel RealSense D435i. We successfully implemented a conversational agent that employs multiple AI models for various tasks, such as face detection or voice analysis. Our work underscores the importance of modular and open hardware design in robotic systems, allowing for easier integration of new components and functionalities without compromising the existing infrastructure beyond the life cycle supported by the manufacturer.

{\addonname}'s applications are multifaceted, ranging from educational settings where it can serve as a versatile tool for teaching programming and robotics to research environments where its adaptability can allow for its future use in HRI and HRC research. Moreover, the potential for {\addonname} to facilitate research in human-robot interaction is significant, given the Pepper robot's friendly appearance and the add-on's ability to seamlessly integrate new sensors and actuators.

In conclusion, this project makes a substantial contribution to the field of social robotics by introducing an open-hardware add-on for the Pepper robot. We expect it will greatly benefit many practitioners who rely on it. {\addonname} not only simplifies software development for the robot but also paves the way for a wide range of applications across different sectors, keeping the platform relevant and productive for years to come.

Future research directions based on this project include exploring the integration of more advanced AI algorithms with {\addonname}, enabling the Pepper robot to perform complex tasks that require learning and adaptation. Additionally, investigating the use of {\addonname} in real-world scenarios such as healthcare, education, and customer service would provide valuable insights into its practical applications and potential impact on society.

\backmatter
\section*{Declarations}

\bmhead{Funding} This research was partially funded by the Niedersächsisches Ministerium für Wissenschaft und Kultur via the Volkswagen Foundation under the Programme \href{https://zukunft.niedersachsen.de/foerderangebot/forschungskooperation-niedersachsen-israel/}{zukunft.niedersachsen: Forschungskooperation Niedersachsen -- Israel} project No.\ 15-76251-5616/2023 (\href{https://nicolas-navarro-guerrero.github.io/projects/romeo/}{ROMEO}) and the \href{https://cis.ieee.org/activities/educational-activites/research-grants}{IEEE Computational Intelligence Society Graduate Student Research Grants}. Open Access funding provided by the Projekt DEAL (Open access agreement for Germany).
\bmhead{Conflict of interest/Competing interests} The authors declare that they have no conflict of interest.
\bmhead{Ethics approval} This article does not contain any studies with human participants or animals performed by any of the authors.

\bmhead{Open Access} This article is licensed under a Creative Commons Attribution 4.0 International License, which permits use, sharing, adaptation, distribution and reproduction in any medium or format, as long as you give appropriate credit to the original author(s) and the source, provide a link to the Creative Commons licence, and indicate if changes were made. The images or other third party material in this article are included in the article's Creative Commons licence, unless indicated otherwise in a credit line to the material. If material is not included in the article's Creative Commons licence and your intended use is not permitted by statutory regulation or exceeds the permitted use, you will need to obtain permission directly from the copyright holder. To view a copy of this licence, visit \url{http://creativecommons.org/licenses/by/4.0/}.

\bmhead{Consent to participate} Not applicable
\bmhead{Consent for publication/Informed consent} Not applicable
\bmhead{Availability of data and materials} CAD Files and additional information are available at \url{https://github.com/mfkuhlmann/Pebre}.
\bmhead{Code availability} Code is available at \url{https://github.com/mfkuhlmann/Pebre}


\bibliography{references}

\end{document}